\documentclass[conference]{IEEEtran}
\usepackage{times}

\usepackage[numbers]{natbib}
\usepackage{multicol}
\usepackage[bookmarks=true]{hyperref}
\usepackage{graphicx}

\usepackage{multirow}
\usepackage{multicol}
\usepackage{graphicx}
\usepackage{xspace}
\usepackage{xcolor}
\usepackage{caption}
\usepackage{wrapfig}
\usepackage{bbding}  
\usepackage{pifont}  

\usepackage{booktabs}
\usepackage{float}
\usepackage{amsmath}  
\usepackage{amssymb}  
\usepackage{bm}
\usepackage{dsfont}
\newcommand{\ours}[0]{\textsc{HoST}\xspace}

\usepackage{hyperref}
\usepackage[capitalise, nameinlink]{cleveref}

\newcommand{\paragraphbegin}[1]{\vspace{0.01in}\noindent\textbf{#1}}

\usepackage{colortbl}
\definecolor{ourcolor}{HTML}{99e0eb}
\definecolor{ourblue}{HTML}{27a2c3}

\definecolor{tablecolor}{HTML}{ccf2f5} 

\definecolor{tablecolor2}{HTML}{ffcdb4}
\definecolor{citecolor}{HTML}{fe7b5b}
\definecolor{grey}{rgb}{0.9, 0.9, 0.9}
\usepackage{amssymb}

\usepackage{listings}
\definecolor{gred}{rgb}{0.859,0.267,0.216}
\definecolor{ggreen}{rgb}{0.059,0.616,0.345}

\definecolor{deepblue}{HTML}{27a2c3}

\definecolor{deepred}{HTML}{fe7b5b}

\newcommand{\ie}[0]{\textit{i.e.},\xspace}

\usepackage[font=small,labelfont=bf]{caption}

\usepackage[font=footnotesize,labelfont=bf]{caption}

\definecolor{citecolor}{HTML}{faa700} 
\definecolor{lblue}{HTML}{ffb114} 
\definecolor{ogreen}{HTML}{2E7D32}
\definecolor{bred}{HTML}{BF360C}
\definecolor{newbrown}{HTML}{795548}

\hypersetup{
    colorlinks=true,
    linkcolor=brown,
    filecolor=magenta,      
    urlcolor=brown,
    citecolor=brown,
}

\newcommand{\ourrow}{\rowcolor{gray!7}}

\newcommand{\ci}[1]{\tiny{\textcolor{gray}{~($\pm #1$})}}

\usepackage{multicol}
\usepackage{multirow}
\usepackage{colortbl}
\usepackage{booktabs}   
\usepackage{bbding} 
\usepackage{graphicx}
\let\labelindent\relax 
\usepackage{enumitem}

\usepackage{bbding}

\usepackage{pifont}
\usepackage{xfrac}
\usepackage{arydshln}

\begin{document}

\title{Learning Humanoid Standing-up Control across \\ Diverse Postures}

\author{\authorblockN{Tao Huang\textsuperscript{2,1} \quad Junli Ren\textsuperscript{1,3}
\quad Huayi Wang\textsuperscript{1,2}
\quad Zirui Wang\textsuperscript{1,4} \quad Qingwei Ben\textsuperscript{1,5} \quad Muning Wen\textsuperscript{1,2} \\ \quad Xiao Chen\textsuperscript{1,5} \quad Jianan Li\textsuperscript{5} \quad Jiangmiao Pang\textsuperscript{1}
}
\authorblockA{
\textsuperscript{1}Shanghai AI Laboratory \quad \textsuperscript{2}Shanghai Jiao Tong University \quad 
\textsuperscript{3}The University of Hong Kong \quad  \\
\textsuperscript{4}Zhejiang University \quad 
\textsuperscript{5}The Chinese University of Hong Kong \\
Website: \href{https://taohuang13.github.io/humanoid-standingup.github.io/}{\texttt{humanoid-standingup.github.io}} \quad Code: \href{https://github.com/OpenRobotLab/HoST}{\texttt{https://github.com/OpenRobotLab/HoST}}
}
}



%

\twocolumn[{%
\renewcommand\twocolumn[1][]{#1}%
\maketitle
\vspace{-0.45cm}
\begin{center}
    \centering
    \captionsetup{type=figure}
     \includegraphics[width=1.0\textwidth]{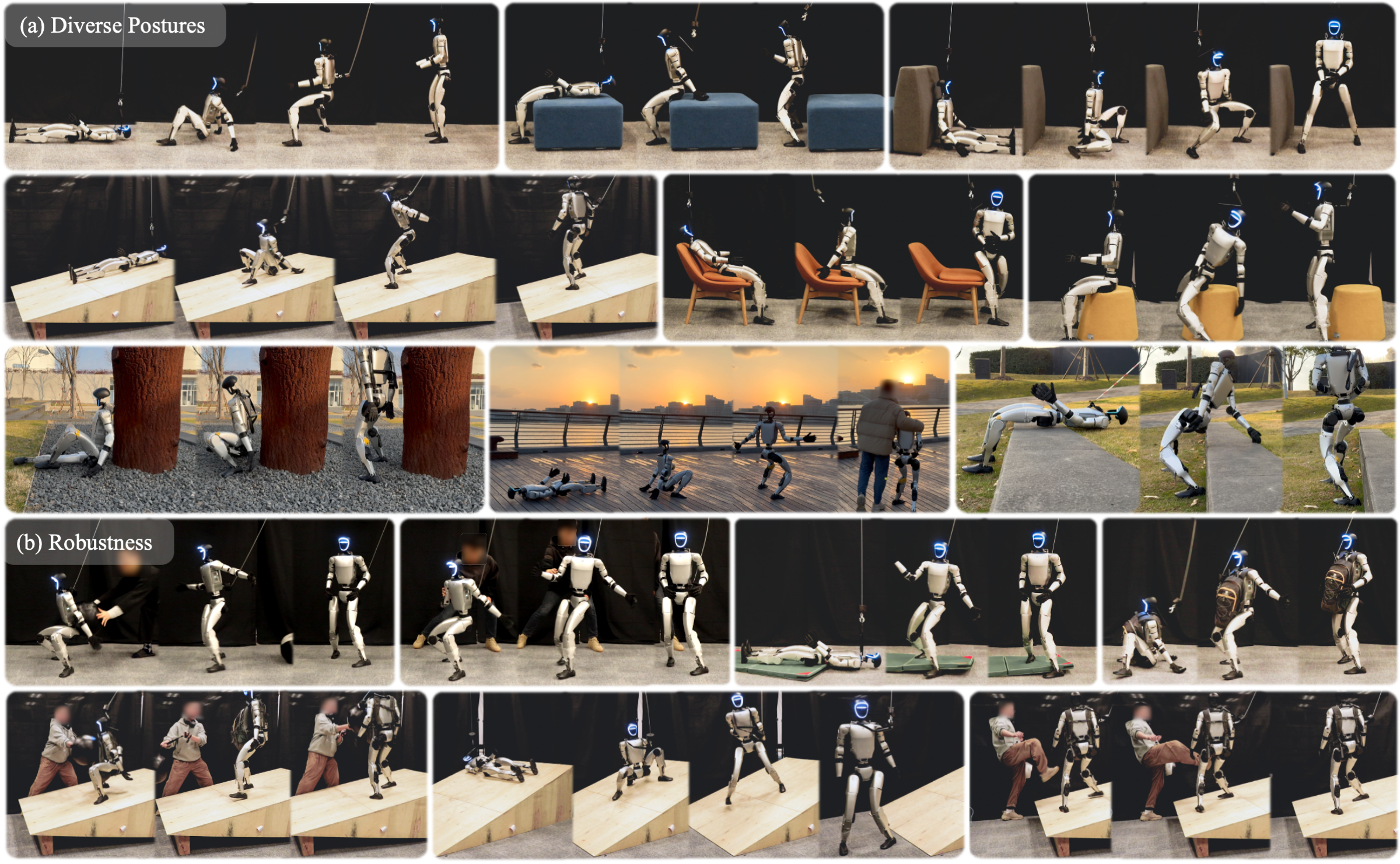}
     \vspace{-0.17in}
    \caption{\textbf{Overview.} (a) Our proposed framework \ours enables the humanoid robot to learn standing-up control via reinforcement learning without prior data, where the robot can successfully stand up across diverse postures in both laboratory and outdoor environments. (b) \ours also demonstrates strong robustness to many environmental disturbances, including external forces, stumbling blocks, 12kg payload, and challenging initial postures.} 
    \label{fig:teaser}
\end{center}
\vspace{0.04in}
}]

\begin{abstract}
Standing-up control is crucial for humanoid robots, 
with the potential for integration into current locomotion and loco-manipulation systems, such as fall recovery. 
Existing approaches are either limited to simulations that overlook hardware constraints or rely on predefined ground-specific motion trajectories, failing to enable standing up across postures in real-world scenes. 
To bridge this gap, we present \ours (\underline{H}uman\underline{o}id \underline{St}anding-up Control), a reinforcement learning framework that learns standing-up control from scratch, enabling robust sim-to-real transfer across diverse postures.
\ours effectively learns posture-adaptive motions by leveraging a multi-critic architecture and curriculum-based training on diverse simulated terrains.
To ensure successful real-world deployment, we constrain the motion with smoothness regularization and implicit motion speed bound to alleviate oscillatory and violent motions on physical hardware, respectively. 
After simulation-based training, the learned control policies are directly deployed on the Unitree G1 humanoid robot. Our experimental results demonstrate that the controllers achieve smooth, stable, and robust standing-up motions across a wide range of laboratory and outdoor environments (\cref{fig:teaser}). Videos and code are available on \href{https://taohuang13.github.io/humanoid-standingup.github.io/}{our project page}.
\vspace{0.in}
\end{abstract}

\IEEEpeerreviewmaketitle

\section{Introduction}
Can humanoid robots stand up from a sofa, walk to a table, and pick up coffee, seamlessly like humans?
Fortunately, recent advancements in humanoid robot hardware and control have enabled significant progress in bipedal locomotion~\cite{radosavovic2024real,li2024reinforcement,long2024learning,zhuang2024humanoid} and bimanual manipulation~\cite{Cheng2024OpenTeleVisionTW,Li2024OKAMITH,Fu2024HumanPlusHS,Jiang2024HarmonWM}, allowing robots to navigate environment and interact with objects effectively. However, the fundamental capability—standing-up control~\cite{stuckler2006getting,kanehiro2003first}—remains underexplored. Most existing systems assume the robots start from a pre-standing posture, limiting their applicability to many scenes, such as transitioning from a seated position or recovering after a loss of balance. We envision that unlocking this standing-up capability would broaden the real-world applications of humanoid robots. To this end, we investigate how humanoid robots can learn to stand up across diverse postures in real environments.

A classical approach for this control task involves tracking handcrafted motion trajectories through model-based motion planning or trajectory optimization~\cite{kanehiro2003first,kanehiro2007getting,kuniyoshi2004dynamic,stuckler2006getting}. 
Although effective in generating motions, these methods require extensive tuning of analytical models and often perform suboptimally in real-world settings with external disturbances~\cite{luo2014multi,lee2019robust} or inaccurate actuator modeling~\cite{hwangbo2019learning}. Besides, real-time optimization on the robot makes these methods computationally intensive, prompting workarounds such as reduced optimization precision or offload computations to external machines~\cite{neunert2017trajectory,farshidian2017efficient}, though both are with practical limitations.

Reinforcement learning (RL) offers an alternative effective framework for humanoid locomotion and whole-body control~\cite{Peng2022ASE,He2024OmniH2OUA,Cheng2024ExpressiveWC,Zhang2024WoCoCoLW}, benefiting from minimal modeling assumptions. However, compared to these tasks that partially decouple upper- and lower-body dynamics, RL-based standing-up control involves a highly dynamic and synergistic maneuver on both halves of the body. This complex maneuver features time-varying contact points~\cite{kanehiro2003first}, multi-stage motor skills~\cite{luo2014multi}, and precise angular momentum control~\cite{Goswami2004RateOC}, making RL exploration challenging. Although predefined motion trajectories can guide RL exploration, they are typically limited to ground-specific postures~\cite{peng2018deepmimic,Peng2022ASE,yang2023learning,haarnoja2024learning}, leaving the scalability to other postures unclear. Conversely, training RL agents from scratch with wide explorative strategies on the ground can lead to violent and abrupt motions that hinder real-world deployment~\cite{tao2022learning}, particularly for robots with many actuators and wide joint limits. In summary, learning posture-adaptive, real-world deployable standing-up control with RL remains an open problem (see \cref{table:comparision_method}).

In this work, we address this problem by proposing \ours, an RL-based framework that learns humanoid standing-up control across diverse postures from scratch. To enable posture-adaptive motion beyond the ground, we introduce multiple terrains for training and a vertical pull force during the initial stages to facilitate exploration. Given the multiple stages of the task, we adopt multi-critic RL~\cite{mysore2022multi} to optimize distinct reward groups independently for a better reward balance. To ensure real-world deployment, we apply smoothness regularization and motion speed constraints to mitigate oscillatory and violent motions. Our control policies, trained in simulation~\cite{makoviychuk2021isaac} with domain randomization~\cite{tobin2017domain}, can be directly deployed on the Unitree G1 humanoid robot. The resulting motions, tested in both laboratory and outdoor environments, demonstrate high smoothness, stability, and robustness to external disturbances, including forces, stumbling blocks, and heavy payloads. 

We overview the real-world performance of our controllers in \cref{fig:teaser} and summarize our core contributions as follows:
\begin{itemize}[leftmargin=4mm]
\vspace{-0.02in}
    \item \textbf{Real-world posture-adaptive motions} are well achieved through our proposed RL-based method, without relying on predefined trajectories or sim-to-real adaptation techniques.
    \item \textbf{Smoothness, stability, and robustness}  are consistently demonstrated by our learned control policies, even under challenging external disturbances.
    \item \textbf{Evaluation protocols} are elaborately designed to analyze standing-up control comprehensively, aiming to guide future research and development in this control task.
\vspace{-0.02in}
\end{itemize}

\begin{table}[t]
    \centering
    \vspace{0.07in}
    \caption{Comparison with existing methods on standing-up control.}
    \resizebox{1\linewidth}{!}{
    \begin{tabular}{lcc ccc}
    \toprule 
    \multirow{2}{*}{Method} & \multirow{2}{*}{\shortstack{Real \\ Robot}} & \multirow{2}{*}{\shortstack{w/o Prior \\ Trajectory }} & \multirow{2}{*}{\shortstack{Beyond \\ Ground}} & \multirow{2}{*}{\shortstack{High \\ DoF}} & \multirow{2}{*}{\shortstack{1-stage \\ Training}} \\ \\
    \midrule
    \citet{Peng2022ASE} & \XSolidBrush & \XSolidBrush & \XSolidBrush & \Checkmark & \XSolidBrush \\
    \citet{yang2023learning} & \XSolidBrush & \XSolidBrush & \XSolidBrush &  \Checkmark &  \Checkmark \\
    \citet{tao2022learning} & \XSolidBrush & \Checkmark & \XSolidBrush & \Checkmark & \XSolidBrush \\ 
    \citet{haarnoja2024learning} & \Checkmark & \XSolidBrush & \XSolidBrush & \Checkmark & \Checkmark \\ 
    \citet{gaspard2024frasa} & \Checkmark & \Checkmark & \XSolidBrush & \XSolidBrush & \Checkmark \\ 
    \cmidrule(r){1-6}
    \ours (ours) & \Checkmark & \Checkmark & \Checkmark & \Checkmark & \Checkmark \\

    \bottomrule
    \end{tabular}}
    \vspace{-0.1in}
    \label{table:comparision_method}
\end{table}

\section{Related Work}
\subsection{Learning Humanoid Standing-up Control}
Classical approaches to standing-up control rely on tracking handcrafted motion trajectories through model-based optimization~\cite{kanehiro2003first,kanehiro2007getting,kuniyoshi2004dynamic,stuckler2006getting}. While effective, these methods are computationally intensive, sensitive to disturbances~\cite{luo2014multi,lee2019robust}, and require precise actuator modeling~\cite{hwangbo2019learning}, limiting their real-world applicability. In contrast, RL-based methods learn control policies with minimal modeling assumptions, either by leveraging predefined motion trajectories to guide exploration~\cite{peng2018deepmimic,Peng2022ASE,yang2023learning,haarnoja2024learning} or employing exploratory strategies to learn from scratch~\cite{tao2022learning}.  However, none of these methods have demonstrated real-world standing-up motion across diverse postures. Our proposed RL framework addresses these limitations by achieving posture adaptivity and real-world deployability without predefined motions, enabling smooth, stable, and robust standing-up across a wide range of laboratory and outdoor environments.

\subsection{Reinforcement Learning for Humanoid Control}
Reinforcement learning (RL) has been effectively applied to various humanoid control tasks, showcasing its versatility and effectiveness. For example, RL has enabled humanoid robots to achieve robust locomotion on diverse terrains~\cite{radosavovic2024real,li2024reinforcement,zhuang2024humanoid,long2024learning}, whole-body control for expressive human-like motions~\cite{peng2018deepmimic,Peng2022ASE,He2024OmniH2OUA,he2024hover,Cheng2024ExpressiveWC}, versatile jumping~\cite{Zhang2024WoCoCoLW}, and loco-manipulation~\cite{dao2024sim,liu2024opt2skill,wang2024hypermotion}. Building on these advances, we address humanoid standing-up control, a parallel problem presenting unique challenges due to its dynamic nature and the need for precise coordination of multi-stage motor skills and time-varying contact points~\cite{kanehiro2003first,luo2014multi}. We propose a novel approach that integrates a multi-critic framework, motion constraints, and a training curriculum to facilitate real-world deployment, setting it apart from prior methods.

\subsection{Learning Quadrupedal Robot Standing-up Control}
Standing-up control for quadrupedal robots shares similarities with humanoid robots but faces distinct challenges due to morphological differences, such as quadrupedal designs. Classical approaches for quadrupedal robots often rely on model-based optimization and predefined motion primitives~\cite{castano2019design,saranli2004model}, which work well in controlled environments but struggle with adaptability to diverse postures and real-world uncertainties. Recent RL-based methods have enabled quadrupedal robots to recover from falls and transition between poses~\cite{lee2019robust,ma2023learning,yang2023learning}, using exploratory learning to manage complex dynamics and environmental interactions. Our work draws inspiration from these advances, extending them to humanoid robots by addressing the unique challenges of bipedal standing-up control. By incorporating posture adaptivity, motion constraints, and a structured training curriculum, our framework bridges the gap between quadrupedal and humanoid robot control, enabling robust standing-up motions across diverse environments.

\section{Problem Formulation}
We formulate the problem of humanoid standing up as a Markov decision process (MDP; \cite{puterman2014markov}) with finite horizon, which is defined by the tuple $\mathcal{M} = \langle \mathcal{S}, \mathcal{A}, \mathcal{T}, \mathcal{R}, \gamma \rangle$.  
At each timestep $t$, the agent (\ie the robot) perceives the state $s_t\in\mathcal{S}$ from the environment and executes an action $a_t\in\mathcal{A}$ produced by its policy $\pi_\theta(\cdot|s_t)$. The agent then observes a successor state $s_{t+1}\sim \mathcal{T}(\cdot|s_t, a_t)$ following the environment transition function $\mathcal{T}$ and receives a reward signal $r_t\in\mathcal{R}$. To solve the MDP, we employ reinforcement learning (RL;~\cite{sutton2018reinforcement}), whose goal learn an optimal policy $\pi_\theta$ that maximizes the expected cumulative reward (return) $\mathbb{E}_{\pi_\theta}[\sum_{t=0}^{T-1}\gamma^t r_t]$ the agent receives during the whole $T$-length episode, where $\gamma\in[0, 1]$ is the discount factor. The expected return is estimated by a value function (critic) $V_\phi$. In this paper, we adopt Proximal Policy Optimization (PPO;~\cite{schulman2017proximal}) as our RL algorithm because of its stability and efficiency in large-scale parallel training. 


\begin{figure}[t]
    \centering
    \includegraphics[width=1\linewidth]{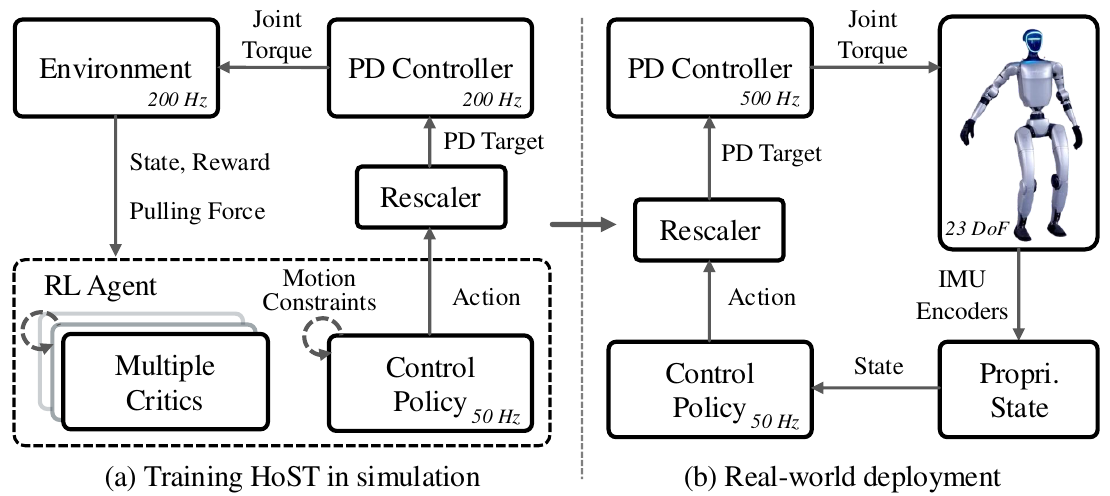}
    \caption{\textbf{Framework overview}. (a) We train policies in simulation from scratch with multiple critics and motion constraints operationalized by rewards, smoothness regularization, and action bound (rescaler). (b) The trained polices can be directly deployed in the real robot to produce standing-up motions.}
    \label{fig:framework}
    \vspace{-0.05in}
\end{figure}

\subsubsection{Observation Space} We hypothesize that the proprioceptive states of robots provide sufficient information for standing-up control in our target environments.
We thus include the proprioceptive information read from robot's Inertial Measurement Unit (IMU) and joint encoders into the state $s_t=[\omega_t, r_t,q_t,p_t, \dot{p}_t, a_{t-1}, \beta]$, where $\omega_t$ is the angular velocity of robot base,  $r_t$ and $q_t$ are the roll and pitch, $p_t$ and $\dot{p}_t$ are positions and velocities of the joints, $a_{t-1}$ is the last action, and $\beta\in (0, 1]$ is a scalar that scale the output action. 
Given the contact-rich nature of the standing-up task, we implicitly enhance contact detection by feeding the policy with the previous five states~\cite{hwangbo2019learning}. 

\subsubsection{Action Space} We employ a PD controller for torque-based robot actuation. The action $a_t$ represents the difference between the current and next-step joint positions, with the PD target computed as $p_t^d = p_t + \beta a_t$, where each dimension of $a_t$ is constrained to $[-1, 1]$. The action rescalar $\beta$ restricts the action bounds to regulate the motion speed implicitly. This is essential to constrain the standing-up motion and will be discussed in later sections. The torque at timestep $t$ is computed as:
\begin{equation}\label{eq:pd}
    \tau_t = K_p\cdot(p_t^d - p_t) - K_d\cdot \dot{p}_t,
\end{equation}
where $K_p$ and $K_d$ represent the stiffness and damping coefficients of the PD controller. The dimension of action space $|A|$ corresponds to the number of robot actuators.

\section{Method}
This section introduces \ours (\underline{H}uman\underline{o}id \underline{St}anding-up Control), a reinforcement learning (RL)-based framework for learning humanoid robots to stand up across diverse postures, as summarized in \cref{fig:framework}. This control task is highly dynamic, multi-stage, and contact-rich, posing challenges for conventional RL approaches. We first outline the key challenges addressed in this work in~\cref{subsec:key_challenges}, then describe the core components of the framework in the following sections.

\subsection{Key Challenges \& Overview}\label{subsec:key_challenges}
\subsubsection{Reward Design \& Optimization (\cref{subsec:multi_critic})} The standing-up task involves multiple motor skills: righting the body, kneeling, and rising. Learning a control policy for these stages is challenging without explicit stage separation~\cite{li2023robust,kim2024stage}. We address this by dividing the task into three stages and activating corresponding reward functions at each stage. The complexity of these skills requires multiple reward functions, which can complicate policy optimization. To mitigate this, we employ multi-critic RL~\cite{mysore2022multi}, grouping reward functions to balance objectives effectively.

\subsubsection{Exploration Challenges (\cref{subsec:force})} Despite multi-critic RL, exploration remains difficult due to the robot’s high degrees of freedom and wide joint limits. Drawing inspiration from human infant skill development~\cite{claxton2012control,von1982eye}, we facilitate exploration by applying a curriculum-based vertical pulling force on the base link of the robot.

\subsubsection{Motion Constraints (\cref{subsec:motion_smoothness})} With only reward functions, the agent tends to learn violent and jerky motions, driven by high torque limits and numerous actuators. Such behaviors are impractical for real-world deployment. To address this, we introduce an action rescaler $\beta$ to gradually tighten action output bounds, implicitly limiting joint torques and motion speed. Additionally, we incorporate smoothness regularization~\cite{Kobayashi2022L2C2LL} to mitigate motion oscillation.

\subsubsection{Sim-to-Real Gap (\cref{subsec:simulation})} A significant challenge is the sim-to-real gap. We address this through two strategies: (1) designing diverse terrains to better simulate real-world starting postures, and (2) applying domain randomization~\cite{tobin2017domain} to reduce the influence of physical discrepancies between simulation and the real world.

\subsection{Reward Functions \& Multiple Critics}\label{subsec:multi_critic}
Considering the multi-stage nature of the task, we divide the task into three stages: righting the body $h_\mathrm{base}<H_{\mathrm{stage1}}$, rising the body $H_{\mathrm{stage1}}<h_\mathrm{base}<H_{\mathrm{stage2}}$, and standing $h_\mathrm{base}>H_{\mathrm{stage2}}$, indicated by the height of the robot base $h_\mathrm{base}$. Corresponding reward functions are activated at each stage. We then classify reward functions into four groups: (1) \textbf{task reward} $r^{\mathrm{task}}$ that specifies the high-level task objectives, (2) \textbf{style reward} $r^{\mathrm{style}}$ that shapes the style of standing-up motion, (3)  \textbf{regularization reward} $r^{\mathrm{regu}}$ that further regularizes the motionw, and (4) \textbf{post-task reward} $r^{\mathrm{post}}$ that specify the desired behaviors after successful standing up, \ie stay standing. The overall reward function is expressed as follows:
\begin{equation*}
    r_t = w^{\mathrm{task}}\cdot r^{\mathrm{task}}_t + w^{\mathrm{style}}\cdot r^{\mathrm{style}}_t +  w^{\mathrm{regu}}\cdot r^{\mathrm{regu}}_t + w^{\mathrm{post}}\cdot r^{\mathrm{post}}_t,
\end{equation*}
where $w$ with superscript represents the corresponding reward weight. Each reward group contains multiple reward functions. A comprehensive list of all reward functions and groups is provided in~\cref{table:reward_functions}.

\begin{figure}[t]
    \centering
    \includegraphics[width=1\linewidth]{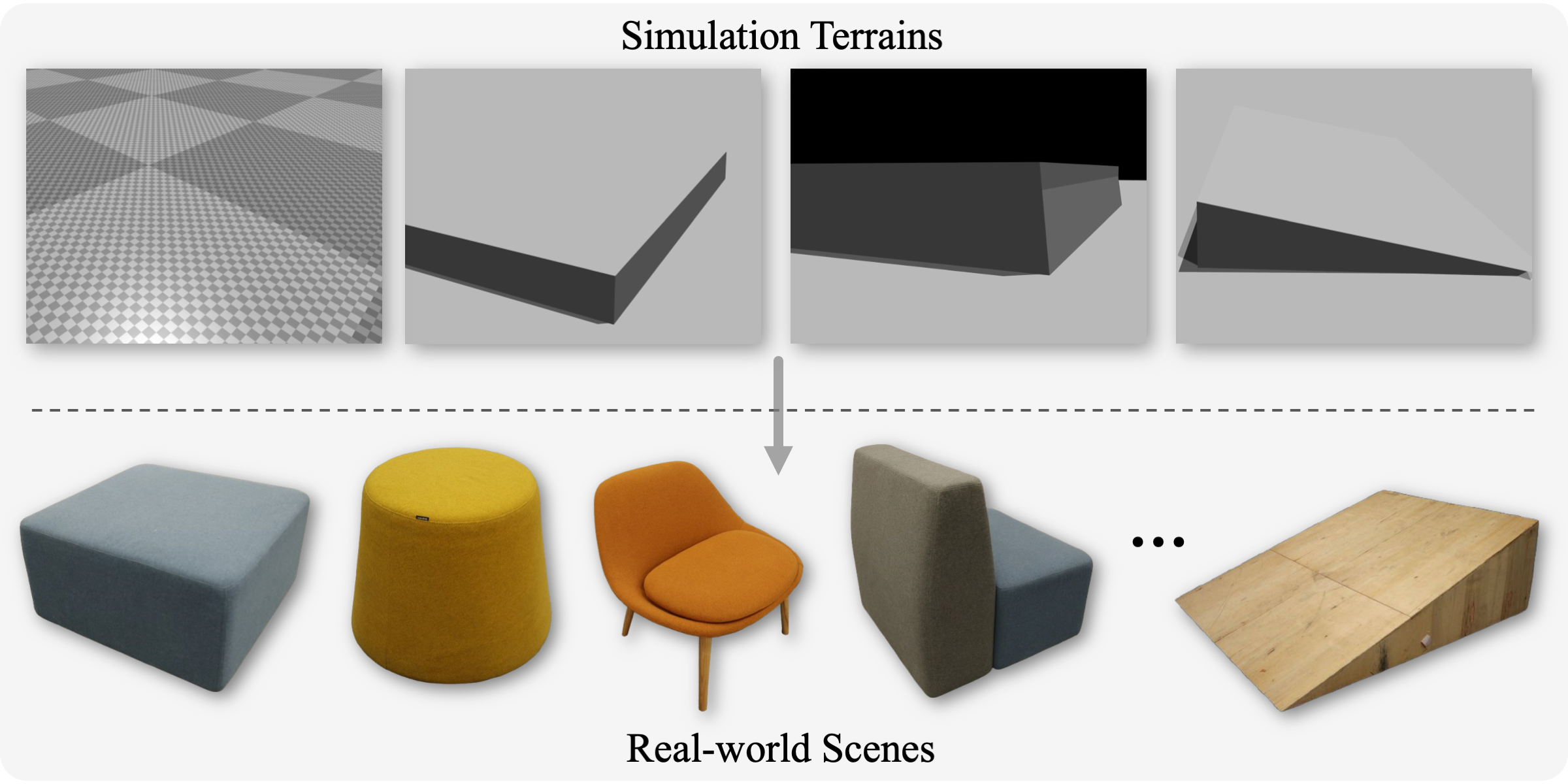}
    \caption{\textbf{Simulation terrains and real-world scenes}. We design four terrains in simulation: ground, platform, wall, and slope to create initial robot postures that are likely to be met in real-world environments. Examples of these real-world environments are shown at the bottom of the figure.}
    \label{fig:terrain_examples}
\end{figure}

However, we observe that using a single value function (critic) presents significant challenges in learning effective standing-up motions. Besides, the large number of reward functions makes hyperparameter tuning computationally intensive and difficult to balance. To address these challenges, we implement multiple critics (MuC;~\cite{mysore2022multi,xu2023composite,zargarbashi2024robotkeyframing}) to estimate returns for each reward group independently, where each reward group is regarded as a separate task with its own assigned critic $V_{\phi_i}$. These multiple critics are then integrated into the PPO framework for optimization as follows:
 \begin{equation}
    \mathcal{L}(\phi_i) = \mathbb{E}\big[ \|r_{t}^i + \gamma V_{\phi_i}(s_t) - \bar{V}_{\phi_i}(s_{t+1})\|^2\big],
\end{equation}
where $r_{t}^i$ is the total reward and $\bar{V}_{\phi_i}$ is the target value function of reward group $i$. Each critic independently computes its advantage function $A_{\phi_i}$ estimated through GAE~\cite{Schulman2015HighDimensionalCC}. These individual advantages are then aggregated into an overall weighted advantage: $A = \sum_{i}  w^i \cdot \frac{A_{\phi_i} - \mu_{A_{\phi_i}}}{\sigma_{A_{\phi_i}}} $, where $\mu_{A_{\phi_i}}$ and $\sigma_{A_{\phi_i}}$ are the batch mean and standard deviation of each advantage. The critics are updated simultaneously with the policy network $\pi_\theta$ according to:
\begin{equation}
    \mathcal{L}(\theta) = \mathbb{E} \left[ \min \left( \alpha_t (\theta)A_t, \mathrm{clip}(\alpha_t(\theta), 1-\epsilon, 1+\epsilon)A_t \right) \right],
\end{equation}
where $\alpha_t(\theta)$ and $\epsilon$ are the probability ratio and the clipping hyperparameter, respectively.

\subsection{Force Curriculum as Exploration Strategy}\label{subsec:force}
 The primary exploration challenges emerge during the transition from falling to stable kneeling, a stage that proves difficult to explore effectively through random action noise alone.
While human infants are likely to learn motor skills with external supports~\cite{claxton2012control,von1982eye}, it inspires us to design environmental assistance to accelerate the exploration. Specifically, we apply an upward force $\mathcal{F}$ on the robot base, which is largely set at the start of training. This force takes effect only when the robot's trunk achieves a near-vertical orientation, indicating a successful ground-sitting posture. The force magnitude decreases progressively as the robot can maintain a target height at the end of the episode. See more details in \cref{app:hyperparams}.

\subsection{Motion Constraints}\label{subsec:motion_smoothness}
\subsubsection{Action Bound (Rescaler)} Humanoid robots often feature many DoFs, each equipped with wide position limits and high-power actuators. This configuration often results in violent motions after RL training, characterized by violent ground hitting and rapid bouncing movements. While setting low action bounds could mitigate this behavior, it might prevent the robot from exploring effective standing-up motions. To this end, we introduce an action rescaler $\beta$ to scale the action output, implicitly controlling the bound of the maximal torques on each actuator. This scale coefficient gradually decreases like vertical force reduction. See more details in \cref{app:hyperparams}.

\begin{table}[t]
    \centering
    \caption{Domain randomization settings for standing-up control.}
    \begin{tabular}{ll}
    \toprule 
    Term & Value \\
    \midrule 
    Trunk Mass & $\mathcal{U}(-2,5)$kg  \\
    Base CoM offset & $\mathcal{U}(-d,d)$m, $d=0.12(XY), 0.08(Z)$ \\
    Link mass & $\mathcal{U}(-0.8,1.2)\times$ default kg  \\ 
    Fiction & $\mathcal{U}(0.1,1)$ \\
    Restitution & $\mathcal{U}(0,1)$ \\
    P Gain & $\mathcal{U}(0.85,1.15)$ \\
    D Gain & $\mathcal{U}(0.85,1.15)$ \\
    Torque RFI~\cite{campanaro2024learning} & $\mathcal{U}(-0.05,0.05)\times$ torque limit N$\cdot$m  \\
    Motor Strength & $\mathcal{U}(0.9,1.1)$ \\
    Control delay & $\mathcal{U}(0, 100)$ms \\
    Initial joint angle offset & $\mathcal{U}(-0.1, 0.1)$rad \\
    Initial joint angle scale & $\mathcal{U}(0.9, 1.1)\times$ default joint angle rad \\
    \bottomrule
    \end{tabular}
    \label{table:domain_randomization}
\end{table}

\begingroup
\setlength{\tabcolsep}{3.5pt}
\begin{table*}[t]
    \centering
    \caption{\textbf{Main simulation results.} We present a performance comparison between \ours and baselines for the proposed metrics. The means and standard variation are reported across 5 evaluations, each with 250 testing episodes. '/' indicates that the method completely failed on a certain task.
    } 
    
    \resizebox{1\linewidth}{!}{%
 \begin{tabular}{lc c cccc c cccc  c cccc c cccc} 
 \toprule
  \multirow{2}{*}{\textbf{Method}} & & \multicolumn{4}{c}{Ground} & & \multicolumn{4}{c}{Platform} & & \multicolumn{4}{c}{Wall} & & \multicolumn{4}{c}{Slope} \\ 
  \cmidrule{3-6}\cmidrule{8-11}\cmidrule{13-16} \cmidrule{18-21} 
    & 
    & $E_{\mathrm{succ}}\uparrow$  & $E_{\mathrm{feet}}\downarrow$ & $E_{\mathrm{smth}}\downarrow$  &$E_{\mathrm{engy}}\downarrow$
    & 
    
    & $E_{\mathrm{succ}}$ $\uparrow$ & $E_{\mathrm{feet}}$ $\uparrow$ & $E_{\mathrm{smth}}$ $\downarrow$ & $E_{\mathrm{engy}}$ $\downarrow$ 
    &
    
    & $E_{\mathrm{succ}}$ $\uparrow$ & $E_{\mathrm{feet}}$ $\uparrow$ & $E_{\mathrm{smth}}$. $\downarrow$ & $E_{\mathrm{engy}}$. $\downarrow$
    &
    
    & $E_{\mathrm{succ}}$ $\uparrow$ & $E_{\mathrm{smth}}$ $\uparrow$ & $E_{\mathrm{smth}}$ $\downarrow$ & $E_{\mathrm{engy}}$ $\downarrow$
    \\ 
 \midrule 
  \ourrow \multicolumn{21}{l}{\textbf{(a) Ablation on Number of Critics}} & \\
  \cdashline{1-21}\noalign{\vskip 0.6mm}
  \ours-w/o-MuC & 
  & 0.0\ci{0.0}  & / & / & / & 
  & 0.0\ci{0.0} & / & / & / & 
  & 0.0\ci{0.0}  & / & / & / &
  & 0.0\ci{0.0}  & / & / & /\\ 
  \ours & 
  & \textbf{99.5\ci{0.4}} & \textbf{1.52\ci{.10}} & \textbf{2.90\ci{.21}} & \textbf{1.35\ci{.02}} & 
  & \textbf{99.8\ci{0.2}} & \textbf{1.16\ci{.04}} & \textbf{3.39\ci{.39}} & \textbf{0.58\ci{.01}} & 
  & \textbf{94.2\ci{1.2}} & \textbf{1.14\ci{.08}} & \textbf{4.66\ci{.69}} & \textbf{1.08\ci{.02}} &
  & \textbf{98.5\ci{0.4}} & \textbf{5.71\ci{.24}} & \textbf{5.31\ci{.45}} & \textbf{0.83\ci{.01}} \\ \cmidrule(r){1-21}
  \ourrow \multicolumn{21}{l}{\textbf{(b) Ablation on Exploration Strategy}} & \\ 
  \cdashline{1-21}\noalign{\vskip 0.6mm}
  \ours-w/o-Force & 
  & 0.0\ci{0.0} & / & / & / & 
  & 6.8\ci{2.0} & 0.12\ci{.02} & 3.39\ci{.40} & 1.98\ci{.02} & 
  & 0.0\ci{0.0} & / & / & / &
  & 0.0\ci{0.0} & / & / & /\\ 
  
  \ours-w/o-Force-RND & 
  & 19.8\ci{1.2} & \textbf{0.87\ci{.11}} & 3.13\ci{.18} & 2.55\ci{.03} & 
  & 99.5\ci{0.4} & 1.66\ci{.11} & 3.55\ci{.37} & 0.78\ci{.01} & 
  & 0.0\ci{0.0} & / & / & / &
  & 0.0\ci{0.0} & / & / & / \\
  
  \ours & 
  & \textbf{99.5\ci{0.4}} & 1.52\ci{0.10} & \textbf{2.90\ci{.21}} & \textbf{1.35\ci{.02}} &
  & \textbf{99.8\ci{0.2}} & \textbf{1.16\ci{.04}} & \textbf{3.39\ci{.39}} & \textbf{0.58\ci{.01}} & 
  & \textbf{94.2\ci{1.2}} & \textbf{1.14\ci{.08}} & \textbf{4.66\ci{.69}} & \textbf{1.08\ci{.02}} &
  & \textbf{98.1\ci{0.4}} & \textbf{5.71\ci{.24}} & \textbf{5.44\ci{.45}} & \textbf{0.89\ci{.01}} \\ \cmidrule(r){1-21}
  \ourrow \multicolumn{21}{l}{\textbf{(c) Ablation on Motion Constraints}} & \\ 
  \cdashline{1-21}\noalign{\vskip 0.6mm}
  \ours-w/o-Bound & 
  & 98.8\ci{0.6} & 7.27\ci{.42} & 9.52\ci{.25} & 3.59\ci{.02} & 
  & 99.4\ci{0.8} & 6.23\ci{.34} & 11.65\ci{.34} & 1.76\ci{.03} & 
  & \textbf{99.6\ci{0.5}} & 5.48\ci{.70} & 8.80\ci{.74} & 1.73\ci{.02} &
  & 82.4\ci{4.4} & 32.22\ci{2.5} & 16.44\ci{.86} & 2.62\ci{.07}\\ 
  
  \ours-Bound0.25 & 
  & \textbf{99.8\ci{0.4}} & \textbf{1.16\ci{.08}} & \textbf{2.75\ci{.19}} & 1.56\ci{.01} &
  & \textbf{99.8\ci{0.1}} & \textbf{0.68\ci{.05}} & \textbf{3.17\ci{.41}} & 0.79\ci{.02} &
  & 84.6\ci{2.5} & \textbf{0.42\ci{.02}} & \textbf{4.23\ci{.71}} & 1.44\ci{.04} &
  & 98.0\ci{1.4} & \textbf{2.74\ci{.16}} & \textbf{4.67\ci{.42}} & 0.90\ci{.02}\\ 
  
  \ours-w/o-L2C2 & 
  & 92.3\ci{0.7} & 2.29\ci{.06} & 4.05\ci{.21} & 1.43\ci{.01} & 
  & \textbf{99.8\ci{0.0}} & 1.93\ci{.07} & 4.47\ci{.42} & 0.92\ci{.02} & 
  & 97.8\ci{1.6} & 1.43\ci{.16} & 5.29\ci{.70} & 1.55\ci{.02} &
  & 98.8\ci{0.8} & 3.93\ci{.24} & 6.32\ci{.46} & 1.12\ci{.02} \\
  
  \ours-w/o-$r^{\mathrm{style}}$ & 
  & 99.2\ci{0.5} & 1.36\ci{.07} & 2.83\ci{.21} & 1.67\ci{.03} & 
  & 82.2\ci{3.5} & 1.18\ci{.08} & 3.56\ci{.40} & 0.67\ci{.03} & 
  & 0.0\ci{0.0} & / & / & / &
  & 21.4\ci{3.2} & 8.61\ci{.12} & 6.49\ci{.54} & 1.69\ci{.05} \\
  
  \ours & 
  & 99.5\ci{0.4} & 1.52\ci{.10} & 2.90\ci{.21} & \textbf{1.35\ci{.02}} &
  & \textbf{99.8\ci{0.2}} & 1.16\ci{.04} & 3.39\ci{.39} & \textbf{0.58\ci{.01}} & 
  & 94.2\ci{1.2} & 1.14\ci{.08} & 4.66\ci{.69} & \textbf{1.08\ci{.02}} &
  & \textbf{98.5\ci{0.4}} & 5.71\ci{.24} & 5.31\ci{.45} & \textbf{0.83\ci{.01}} \\ \cmidrule(r){1-21}
  \ourrow \multicolumn{21}{l}{\textbf{(d) Ablation on Historical States}} & \\ 
  \cdashline{1-22}\noalign{\vskip 0.6mm}
  \ours-History0 & 
  & 98.1\ci{1.4} & 2.11\ci{.14} & 2.72\ci{.22} & 1.27\ci{.02} & 
  & 99.5\ci{0.5} & 1.53\ci{.13} & 3.29\ci{.40} & \textbf{0.47\ci{.01}} & 
  & 64.5\ci{1.2} & 1.66\ci{.04} & 4.74\ci{.72} & 1.66\ci{.03} &
  & 97.4\ci{2.0} & 5.20\ci{.24} & \textbf{4.97\ci{.48}} & \textbf{0.66\ci{.02}}\\ 
  
  \ours-History2 & 
  & 99.3\ci{0.3} & 2.25\ci{.13} & \textbf{2.56\ci{.19}} & \textbf{1.16\ci{.01}} & 
  & 99.4\ci{0.5} & \textbf{0.77\ci{.39}} & \textbf{3.27\ci{.39}} & 0.60\ci{.01} & 
  & 93.7\ci{1.4} & 1.79\ci{.08} & 4.81\ci{.71} & 1.22\ci{.01} &
  & 98.6\ci{0.6} & 5.06\ci{.24} & 5.35\ci{.44} & 0.77\ci{.01} \\
  
  \ours-History5 (ours) & 
  & \textbf{99.5\ci{0.4}} & \textbf{1.52\ci{.10}} & 2.90\ci{.21} & 1.35\ci{.02} & 
  & \textbf{99.8\ci{0.2}} & 1.16\ci{.04} & 3.39\ci{.39} & 0.58\ci{.01} & 
  & \textbf{94.2\ci{1.2}} & \textbf{1.14\ci{.08}} & 4.66\ci{.69} & \textbf{1.08\ci{.02}} &
  & \textbf{98.6\ci{0.4}} & 5.71\ci{.24} & 5.31\ci{.45} & 0.83\ci{.01} \\
  
  \ours-History10 & 
  & 98.8\ci{0.8} & 1.62\ci{.08} & 3.02\ci{.20} & 1.60\ci{.02} & 
  & 99.2\ci{0.8} & 0.78\ci{.05} & 3.55\ci{.40} & 0.71\ci{.01} & 
  & 88.2\ci{2.6} & 1.24\ci{.06} & \textbf{4.61\ci{.72}} & 1.46\ci{.05} &
  & \textbf{98.6\ci{0.8}} & \textbf{3.93\ci{.26}} & 5.41\ci{.49} & 0.91\ci{.01} \\
\bottomrule

\end{tabular}}
\label{table:main_results}
\vspace{-0.4cm}
\end{table*}
\begin{figure*}[t]
    \centering
    \includegraphics[width=0.97\textwidth]{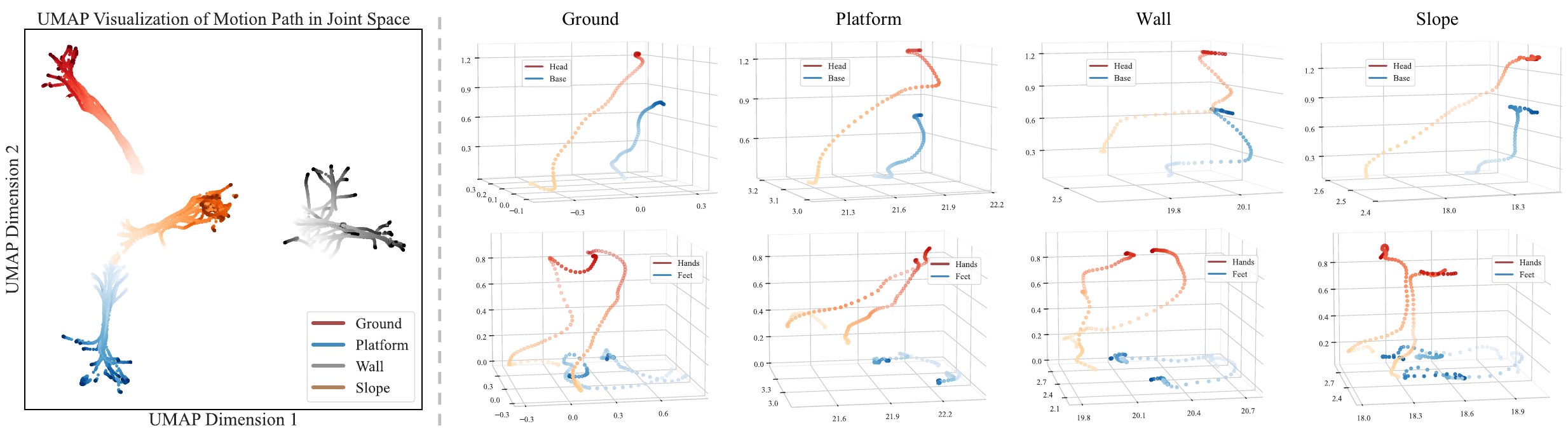}
    \caption{\textbf{Motion analysis in simulation}. (Left) UMAP visualization of joint-space trajectories demonstrates similar but distinct motion patterns on the terrains except for the wall. Besides, the trajectories of each terrain are overall consistent, with variation to handle the difference of starting postures. (Right) 3D trajectory visualizations reveal stable, coordinated hand-foot motion and dynamic posture adaptability, demonstrating effective whole-body coordination and validating the proposed framework. Point color in the plot indicates motion progression, with lighter shades for earlier and darker for later times.}
    \label{fig:trajectory}
    \vspace{-0.1in}
\end{figure*}

\subsubsection{Smoothness Regularization} To prevent motion oscillation, we adopt the smoothness regularization method L2C2~\cite{Kobayashi2022L2C2LL} into our multi-critic formulation. This method applies regularization to both the actor-network $\pi_\theta$ and critics $V_{\phi_i}$ by introducing a bounded sampling distance between consecutive states $s_t$ and $s_{t+1}$:
\begin{equation*}
   \mathcal{L}_{\mathrm{L2C2}} = \lambda_\pi D(\pi_\theta(s_t), \pi_\theta(\bar{s}_t)) +\lambda_V \sum D(V_{\phi_i}(s_t), V_{\phi_i}(\bar{s}_t)),
\end{equation*}
where $D$ is a distance metric, $\lambda_\pi$ and $\lambda_V$ are weight coefficient, $\bar{s}_t=s_t+(s_{t+1}-s_t)\cdot u$ is the interpolated state given a uniform noise $u\sim \mathcal{U}(\cdot)$. We combine this objective function with ordinary PPO objectives to train our control policies. 

\subsection{Training in Simulation \& Sim-to-Real Transfer}\label{subsec:simulation}
We use Isaac Gym~\cite{makoviychuk2021isaac} simulator with 4096 parallel environments and the 23-DoF Unitree G1 robot to train standing-up control policies with the PPO~\cite{schulman2017proximal} algorithm.

\subsubsection{Terrain Design}\label{subsec:terrain} To model the diverse starting postures in the real world, we design 4 terrains to diversify the starting postures: (1) \textbf{ground} that is flat, (2) \textbf{platform} that supports the trunk of robot, (3) \textbf{wall} that supports the trunk of the robot, and (4) \textbf{slope} with a benign inclination that supports the whole robot.  We visualize these terrains and examples of their corresponding scenes in the real world in \cref{fig:terrain_examples}. 

\subsubsection{Domain Randomization}\label{subsec:domain_randomization} To enhance real-world deployment, we employ domain randomization~\cite{tobin2017domain} to bridge the physical gap between simulation and reality. The randomization parameters, detailed in~\cref{table:domain_randomization}, include body mass, base center of mass (CoM) offset, PD gains, torque offset, and initial pose, following~\cite{campanaro2024learning,long2024learning}. Notably, the CoM offset is critical, as it enhances controller robustness against real-world CoM position noise, which may arise from insufficient torques or discrepancies between simulated and real robot models.

\subsection{Implementation Details} Our implementation of PPO is based on \cite{Rudin2021LearningTW}. The actor and critic networks are structured as 3-layer and 2-layer MLPs, respectively. Each episode has a rollout length of 500 steps. For smoothness regularization, the weight coefficients $\lambda_\pi$ and $\lambda_V$ are set to 1 and 0.1, respectively. The PD controller operates at 200 Hz in simulation and 500 Hz on the real robot to ensure accurate tracking of the PD targets, while the control policies run at 50 Hz. Additional implementation details and hardware setup are provided in \cref{app:implementation}.

\begin{figure*}[t]
    \centering
    \includegraphics[width=0.96\textwidth]{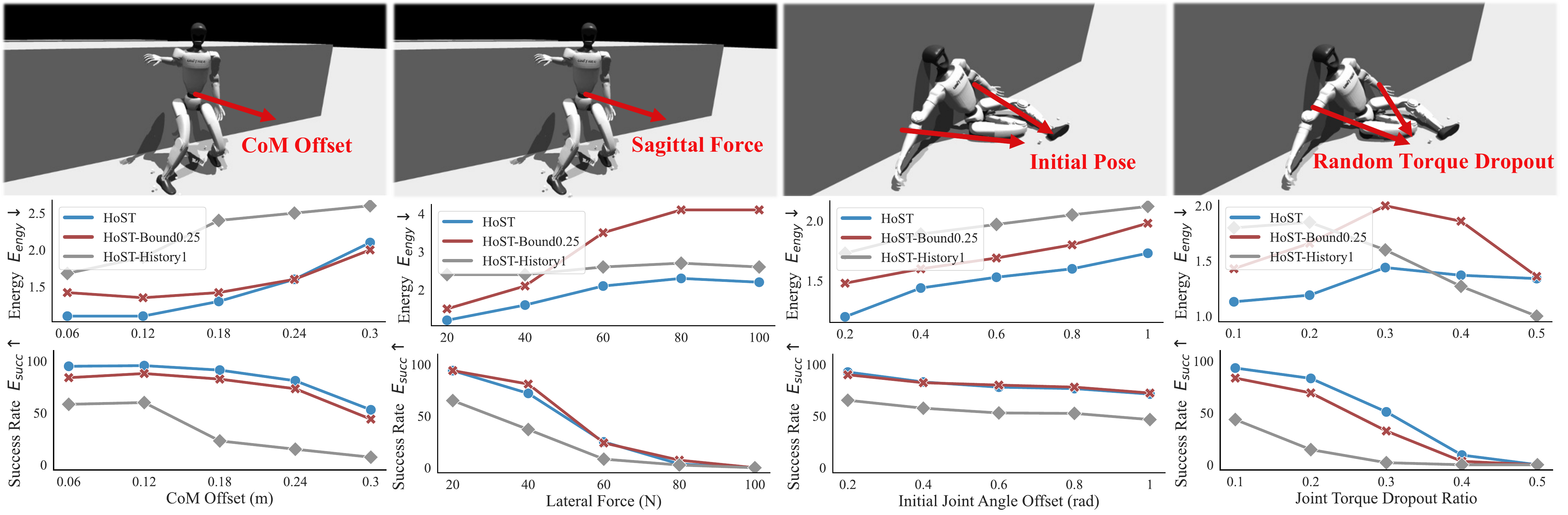}
    \vspace{-0.02in}
    \caption{\textbf{Robustness analysis in simulation}. Evaluation of control policies under four environmental disturbances demonstrates the robustness of our controllers. The poor performance of \ours-History1 indicates the importance of historical information for robustness, while \ours-Bound0.25's high energy consumption reveals limitations in motion quality under disturbance, demonstrating the effect of curriculum setup of action bound.}
    \label{fig:robustness}
    \vspace{-0.08in}
\end{figure*}

\begin{figure}[t]
    \centering
    \includegraphics[width=1\linewidth]{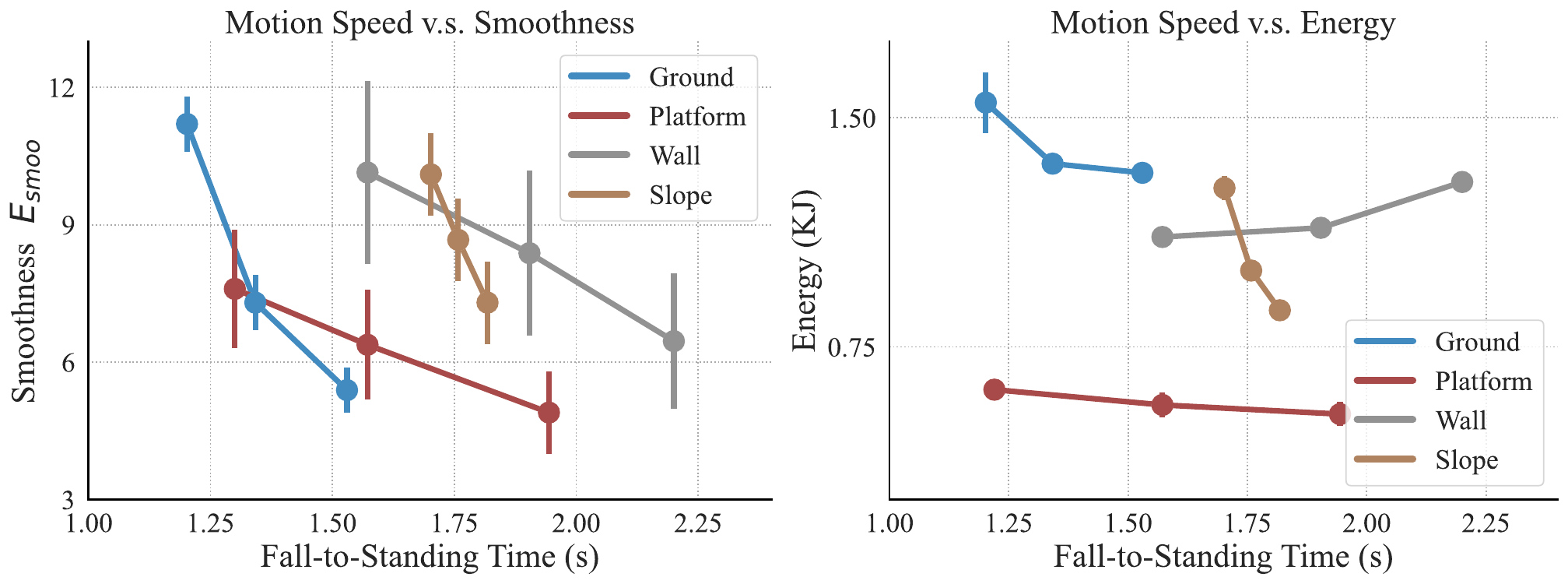}
    \caption{\textbf{Trade-off analysis in simulation}. Trade-offs between motion speed, smoothness, and energy across terrains. Results show the inverse speed-smoothness relationship, indicating the importance of constrained motion speed achieved by our method for real-world deployment.}
    \label{fig:tradeoff}
    \vspace{-0.1in}
\end{figure}

\section{Simulation Experiments}
\subsection{Experimenrt Setup} 
\subsubsection{Evaluation Metrics.} While the design of evaluation metrics for humanoid standing-up control remains an open question~\cite{subburaman2023survey}, we aim to make a step forward by proposing the following metrics: 
\begin{itemize}[leftmargin=4mm]
\vspace{-0.02in}
\item \textbf{Success rate} $E_{succ}$: The episode is considered successful if the robot's base height, $h_{\mathrm{base}}$, exceeds a target height $h_\mathrm{targ}$ and is maintained for the remainder of the episode, indicating stable standing. 
\item \textbf{Feet movement} $E_{\mathrm{feet}}$: The distance traveled by the robot's feet after reaching the target height $h_\mathrm{targ}$, indicating stability in the standing pose. 
\item \textbf{Motion smoothness} $E_{\mathrm{smth}}$: We aggregate the movement of all joint angles of consecutive control steps to measure the smoothness of the motion. It indicates that the robot should keep a smooth motion during the whole episode. 
\item \textbf{Energy} $E_{\mathrm{engy}}$: The energy consumed before reaching $h_\mathrm{targ}$, indicating the avoidance of violent standing-up motion. 
\vspace{-0.02in} 
\end{itemize}

\subsubsection{Baselines} To evaluate the effectiveness of the key design choices in \ours, we compare it against the following ablated versions:
\begin{itemize}[leftmargin=4mm]
\vspace{-0.02in}
    \item \textbf{Single critic}: A baseline using a single critic RL to assess the impact of multiple critics on motor skill learning. 
    \item \textbf{Exploration strategy}: Baselines with random noise and curiosity-based rewards (e.g., RND~\cite{burda2019exploration}) to evaluate the effectiveness of the force curriculum. 
    \item \textbf{Motion constraints}: Ablation of action bounds $\beta$ and smoothness regularization L2C2 to test their influence on motion smoothness.
    \item \textbf{Historical states}: Ablation of the number of historical states to assess their effect on standing-up motion.
\vspace{-0.02in}
\end{itemize}
 
\subsection{Main Results} 
\ours demonstrates good efficacy in learning standing-up control across all terrains, as shown in \cref{table:main_results}. The effect of key design choices is summarized as follows:

\paragraphbegin{Multiple critics are crucial for learning motor skills} Using the same reward functions, the performance of the single critic version of \ours deteriorates significantly across all terrains, achieving zero success rates. This highlights the importance of multiple critics in learning and integrating motor skills while also reducing the hyperparameter tuning burden.

\paragraphbegin{Force curriculum enhances exploration efficiency.} Without the proposed force curriculum, the robot fails to stand up on all terrains except the platform, as the other terrains require exploration from a fully fallen state to stable kneeling. While curiosity-based exploration partially alleviates this challenge, performance remains unsatisfactory. In contrast, the force curriculum greatly improves exploration efficiency.

\paragraphbegin{Action bound prevents abrupt motions.}  While the robot can learn to stand up without action bounds (\ours-w/o-Bound), its movements are excessively violent, as indicated by three performance metrics. With action bounds, \ours demonstrates smoother motions and higher success rates. Although \ours-Bound0.25 performs well, its motions are less natural due to restricted exploration during training.

\paragraphbegin{Smoothness regularization prevents motion oscillation.} Adding smoothness constraints significantly reduces motion oscillation and increases energy efficiency, validating the effectiveness of smooth regularization. Further discussion is presented in \cref{sec:real_robot_exp}.

\begin{figure*}[t]
    \centering
    \includegraphics[width=1\textwidth]{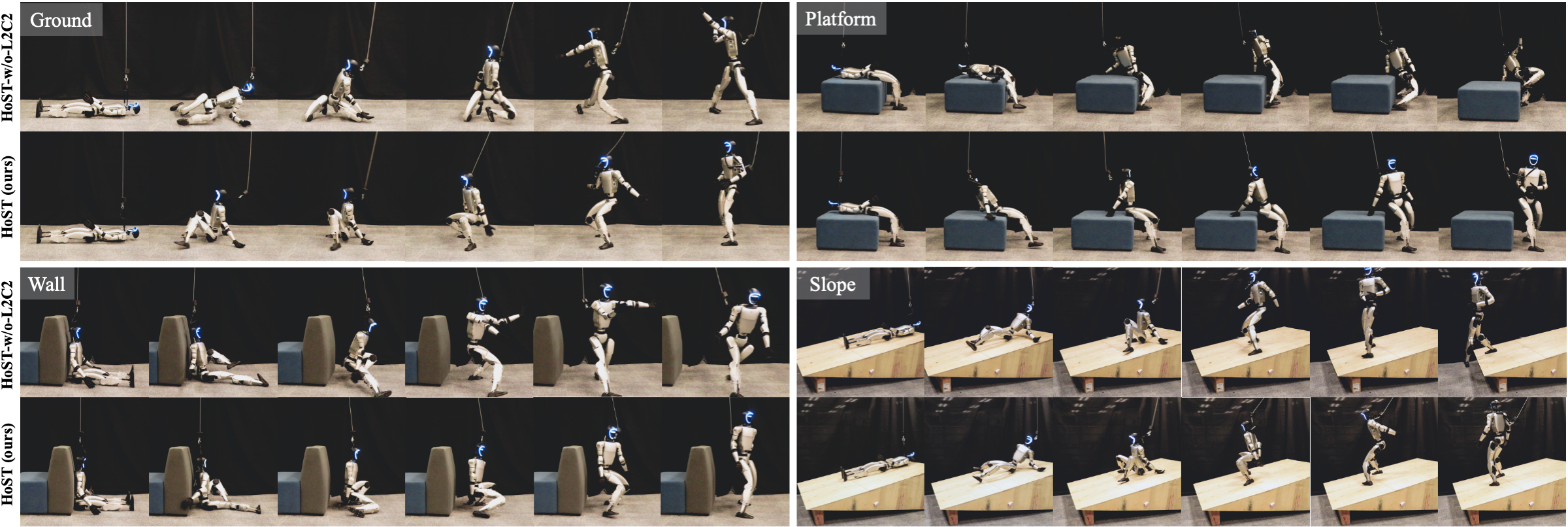}
    \caption{\textbf{Snapshot of real robot motion}. We directly transfer our policies from simulation to four real-world scenes that correspond to four simulation terrains. We conclude that (1) our policies can produce smooth and successful standing-up motion in all tested scenes and (2) smooth regularization of L2C2 is important to avoid oscillation and improve stability.}
    \label{fig:real_snapshot}
    \vspace{-0.05in}
\end{figure*}

\begin{figure*}[t]
    \centering
    \vspace{-0.1in}
    \includegraphics[width=1\textwidth]{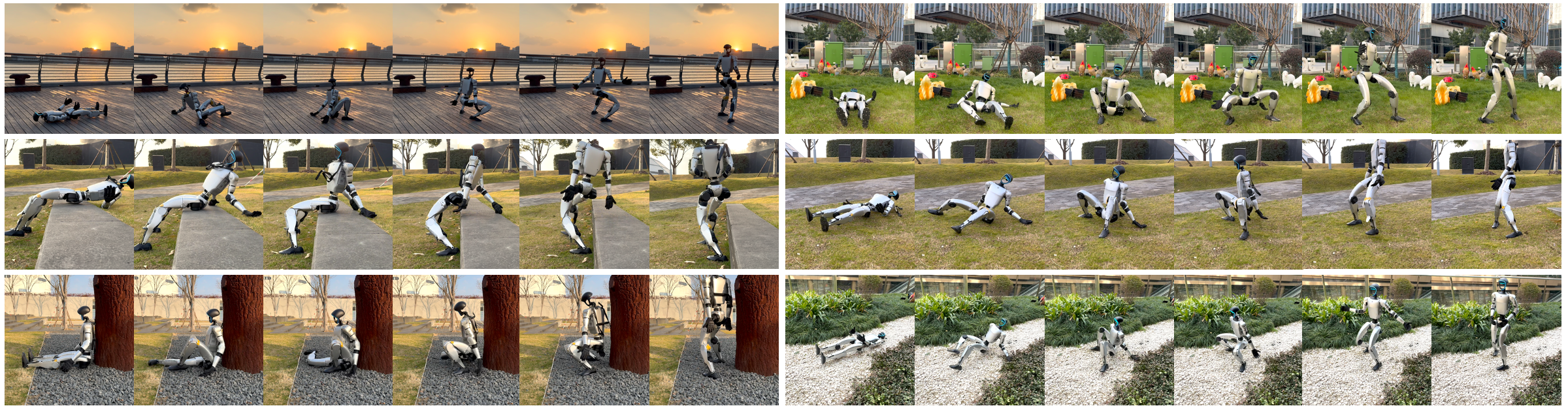}
    \caption{\textbf{Snapshot of outdoor experiments}. We test our controllers in diverse outdoor environments, demonstrating smooth motion on unseen terrains such as grassland, wooden platforms, and stone roads, as well as successful performance on stone platforms and tree-leaning postures.}
    \label{fig:outdoor_snapshot}
\end{figure*}
\begingroup
\setlength{\tabcolsep}{4pt}
\begin{table*}[t]
    \centering
    \vspace{-0.1in} 
    \caption{\textbf{Main results for real robot experiments.} We report the success rate and motion smoothness to quantitatively compare our methods with the baseline. The results demonstrate the superiority of our method and the importance of adding smooth regularization into our method.} 
    
 \begin{tabular}{lc c cc c cc  c cc c cc c cc} 
 \toprule
  \multirow{2}{*}{Method} & & \multicolumn{2}{c}{Ground} & & \multicolumn{2}{c}{Platform} & & \multicolumn{2}{c}{Wall} & & \multicolumn{2}{c}{Slope} & & \multicolumn{2}{c}{Overall} \\ 
   \cmidrule{3-4}\cmidrule{6-7}\cmidrule{9-10} \cmidrule{12-13}  \cmidrule{15-16} 
    & 
    & $E_{\mathrm{succ}}\uparrow$ &  $E_{\mathrm{smth}}\downarrow$ 
    & 
    & $E_{\mathrm{succ}}$ $\uparrow$ & $E_{\mathrm{smth}}$ $\downarrow$ & 
    & $E_{\mathrm{succ}}$ $\uparrow$ & $E_{\mathrm{smth}}$ $\downarrow$ & 
    & $E_{\mathrm{succ}}$ $\uparrow$ &  $E_{\mathrm{smth}}$ $\downarrow$ &
    & $E_{\mathrm{succ}}$ $\uparrow$ &  $E_{\mathrm{smth}}$ $\downarrow$ &
    \\ 
 \midrule 
 \ours-w/o-L2C2 & 
 & \scalebox{1.5}{\sfrac{5}{5}} & 2.09 & 
 & \scalebox{1.5}{\sfrac{2}{5}} & 7.85 & 
 & \scalebox{1.5}{\sfrac{4}{5}} & 13.36 & 
 & \scalebox{1.5}{\sfrac{0}{5}} & 2.89 & 
 & \scalebox{1.5}{\sfrac{11}{20}} & 6.54\\ 
 \ourrow \ours (ours) & 
 & \scalebox{1.5}{\sfrac{5}{5}} & 1.83 & 
 & \scalebox{1.5}{\sfrac{5}{5}} & 5.06 & 
 & \scalebox{1.5}{\sfrac{5}{5}} & 7.22 & 
 & \scalebox{1.5}{\sfrac{5}{5}} & 1.94 & 
 & \scalebox{1.5}{\sfrac{20}{20}} & 4.01 & 
 \\
\bottomrule
\end{tabular}
\label{table:main_real_results}
\vspace{-0.05in}
\end{table*}
\endgroup

\paragraphbegin{Medium history length yields great performance.} \ours with short history length underperforms in contact-rich scenarios, such as the Wall terrain. In contrast, a longer history length improves performance, though it slightly reduces motion smoothness and increases energy consumption compared to the default setting.

\subsection{More Analyses}  
\paragraphbegin{Trajectory analysis (\cref{fig:trajectory}).} Following~\cite{haarnoja2024learning}, we apply Uniform Manifold Approximation and Projection (UMAP;~\cite{mcinnes2018umap}) to project joint-space motion trajectories into 2D, providing a visualization of the humanoid robot’s motion across diverse terrains. The resulting UMAP figure demonstrates distinct motion patterns: smooth, controlled movement on flat ground, while more complex, yet consistent, trajectories emerge on challenging terrains such as Wall. Additionally, in the 3D trajectory plots, the coordinated motion of the robot's hands and feet reveals significant posture adaptability, as the robot adjusts its stance dynamically for balance and stability. These observations highlight the harmonious whole-body coordination achieved by our controllers and validate the effectiveness of our proposed framework.

\paragraphbegin{Robustness analysis (\cref{fig:robustness})}. We comprehensively evaluate the robustness of our learned control policies by simulating various environmental disturbances. Specifically, we test four types of external perturbations: CoM position offset in the sagittal direction, consistent sagittal force,  initial joint angle offset, and random torque dropout ratio. Our results demonstrate that the policies exhibit remarkable robustness across all disturbances, achieving high success rates and efficient motion energy utilization. Notably, the poor performance of \ours-History1 underscores the critical role of historical information, which implicitly encodes contact dynamics, in maintaining robustness. Furthermore, while \ours-Bound0.25 achieves a high success rate, its elevated energy consumption highlights its limited ability to maintain motion smoothness under disturbance. These findings validate the robustness of our policies while indicating the importance of historical context and curriculum of action bound for robust standing-up.

\begin{figure*}[t]
    \centering
    \includegraphics[width=1\textwidth]{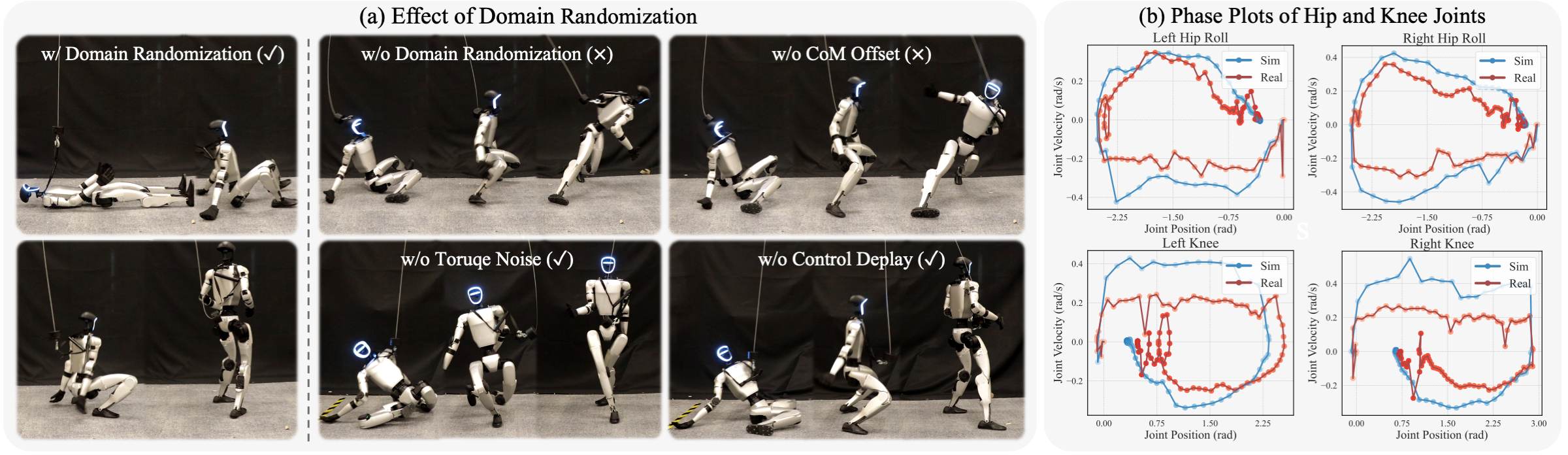}
    \vspace{-0.2in}
    \caption{\textbf{Sim-to-real analysis}. (a) We analyze the effect of each domain randomization term, showing that our randomization terms effectively mitigate the sim-to-real gap, with the CoM position being particularly influential. (b) To further investigate the sim-to-real gap, we compare the phases of knee and hip joints that are crucial for standing-up control. The results reveal significant discrepancies in joint velocities, suggesting a sim-to-real gap in joint torques. }
    \vspace{-0.13in}
    \label{fig:sim2real}
\end{figure*}
\begin{figure*}[t]
    \centering
    \includegraphics[width=1\textwidth]{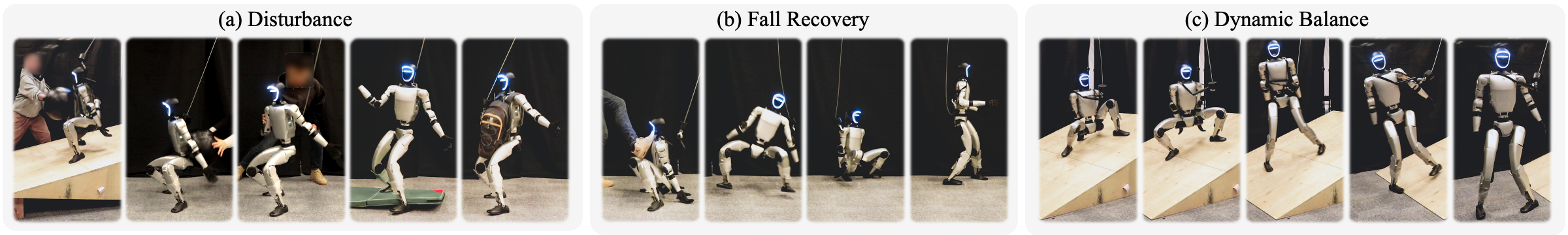}
    \caption{\textbf{Emergent properties in real robot experiments}. (a) our controllers show great robustness to the external force (3kg ball), blocking objects on the ground, and payload mass up to 12kg (2x mass of trunk. (b) Our controllers also exhibit a surprising ability to recover from very large external forces without fully falling down. (c) Our policies also exhibit the ability of dynamic balancing over a 15$^\circ$ slippery slope without falling down.}
    \label{fig:real_robustness}
    \vspace{-0.17in}
\end{figure*}

\paragraphbegin{Trade-off analysis (\cref{fig:tradeoff}).} We examine trade-offs between motion speed, smoothness, and energy consumption across terrains. On the left, motion speed and smoothness exhibit an inverse relationship: longer fall-to-standing times enhance smoothness but reduce speed, a trend consistent across all terrains. On the right, energy consumption increases with fall-to-standing time, with terrain-specific variations. For example, the Slope terrain requires higher energy for balancing. Interestingly, the Wall terrain shows a distinct trend: energy consumption rises sharply at longer fall-to-standing times despite low motion speed, suggesting greater energy intensity. This is likely due to the need for increased force or modified body mechanics to push against a vertical surface, making the motion in Wall less energy-efficient than other terrains. Overall, the results reveal a clear inverse relationship between motion speed and smoothness, indicating the importance of constrained motion speed for real-world deployment and validating the necessity of our approach to achieve such motions.

\begingroup
\setlength{\tabcolsep}{4pt}
\begin{table}[t]
    \centering
    \caption{\textbf{Robustness to payload and random torque dropout.} } 
    
    \resizebox{0.96\linewidth}{!}{%
 \begin{tabular}{lc c ccccc  c ccc c  } 
 \toprule
  \multirow{2}{*}{Metric} & & \multicolumn{5}{c}{Payload Mass} &  &\multicolumn{4}{c}{Torque Dropout Ratio} \\ 
   \cmidrule{3-7} \cmidrule{9-12} 
    & 
    & 4kg &  6kg 
    & 8kg & 10kg 
    & 12kg  & 
    & 0.05 & 0.1 & 0.15 & 0.2
    \\ 
 \midrule 
  $E_{\mathrm{smth}}\downarrow$ & 
  & 1.75 & 1.92 & 1.86 & 1.82 & 1.85 & & 2.00 & 2.16 & 2.61 & / &
 \\[0.4ex]
 $E_{\mathrm{succ}}\uparrow$ & & \scalebox{1.5}{\sfrac{3}{3}} & \scalebox{1.5}{\sfrac{3}{3}} & \scalebox{1.5}{\sfrac{3}{3}} & \scalebox{1.5}{\sfrac{3}{3}} & \scalebox{1.5}{\sfrac{2}{3}} & & \scalebox{1.5}{\sfrac{3}{3}} & \scalebox{1.5}{\sfrac{3}{3}} & \scalebox{1.5}{\sfrac{3}{3}} & \scalebox{1.5}{\sfrac{0}{3}} &
 \\
\bottomrule
\end{tabular}}
\label{table:payload_torque}
\vspace{-0.07in}
\end{table}
\endgroup



\section{Real Robot Experiments}\label{sec:real_robot_exp}

\subsection{Main Results}
We evaluate our method in both laboratory and outdoor environments corresponding to simulation terrains, using \ours-w/o-L2C2 as the baseline to examine the effect of smoothness regularization during deployment.

\paragraphbegin{Smooth regularization improves motions (\cref{fig:real_snapshot}).} Motion oscillations are observed in all scenes without smoothness regularization, often leading to standing-up failures. In contrast, our method produces smooth and stable motions, especially on 10.5$^\circ$ slope. Quantitative results in \cref{table:main_real_results} strengthen this conclusion, with our approach achieving a 100\% success rate and high motion smoothness across all scenes.\footnote{We select the successful episode to compute smoothness to reflect the effect of L2C2 regularization better. Due to the unavailability of the height, we compute the smoothness $E_{\mathrm{smth}}$ within two seconds after starting up. }

\begin{figure}[t]
    \centering
    \includegraphics[width=1\linewidth]{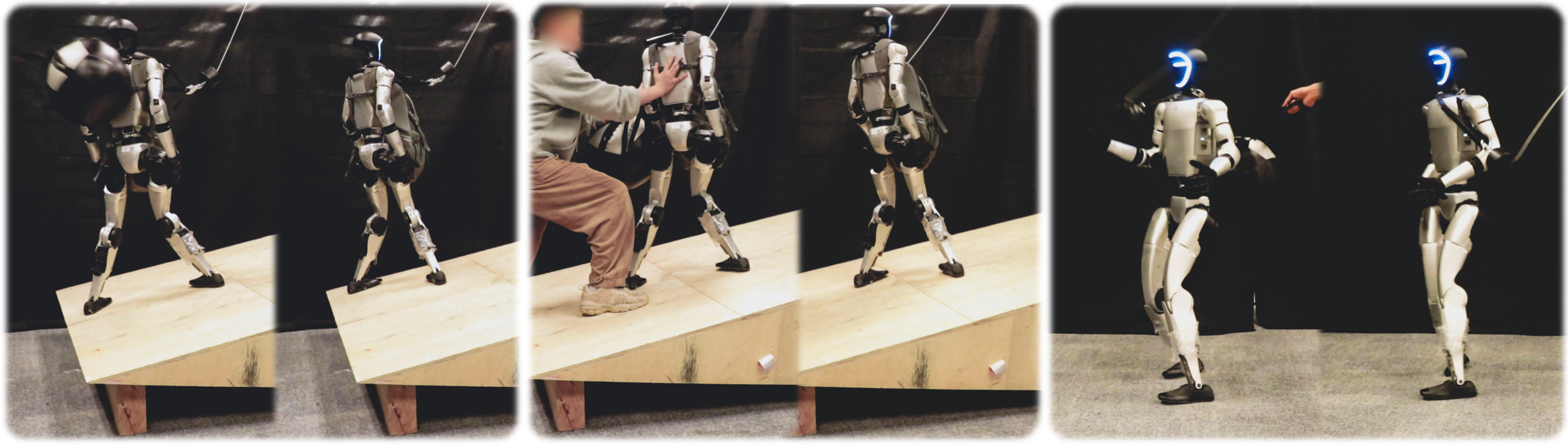}
    \caption{\textbf{Standing stability.} Our control policies demonstrate great stability against external disturbances after successful standing up.}
    \vspace{-0.1in}
    \label{fig:standing_stability}
\end{figure}

\paragraphbegin{Generalization to outdoor environments (\cref{fig:outdoor_snapshot}).} We evaluate our learned controllers in a variety of outdoor environments, testing their ability to generalize to terrains not encountered during training. On flat ground, the controllers produce stable, smooth motions across grassland, wooden platforms, and stone roads. Notably, these terrains were not included in the training simulations. Additionally, our controllers successfully handle more complex scenarios, including stone platforms and tree-leaning postures, demonstrating their adaptability to diverse real-world conditions.

\subsection{Sim-to-real Analysis}\label{subsec:sim2real}
In this analysis, we investigate the effect of various domain randomization terms on the sim-to-real gap, as shown in \cref{fig:sim2real}. Our results demonstrate that the introduction of these randomization terms significantly reduces the sim-to-real gap, particularly with respect to the Center of Mass (CoM) position.

\paragraphbegin{Phase plot.} To further investigate the sources of this gap, we examine the phase plots of the knee and hip roll joints. These joints are considered most important for standing-up motions. We observe a notable discrepancy between simulated and real-world joint velocities, suggesting a gap in joint torques. This highlights the need for more accurate actuator modeling to bridge the sim-to-real gap in humanoid standing-up tasks, which is also suggested by previous work on quadrupedal robots~\cite{hwangbo2019learning}. Despite this, our controllers remain effective in handling these discrepancies, exhibiting joint paths consistent with the simulated ones.

\subsection{Emergent Properties} 

\paragraphbegin{Robustness to external disturbance (\cref{fig:real_robustness}a).} The robustness of our control policies was tested through experiments involving external disturbances, such as a 3 kg ball impact and obstructive objects. The controllers maintained stability even under significant disturbances, like objects disrupting the robot's center of gravity. Additionally, the controllers managed payloads up to 12kg, twice the mass of the humanoid robot's trunk. We also quantitatively verify the great robustness of payload and torque dropout ratio in \cref{table:payload_torque}.  

\begin{figure}[t]
    \centering
    \includegraphics[width=1\linewidth]{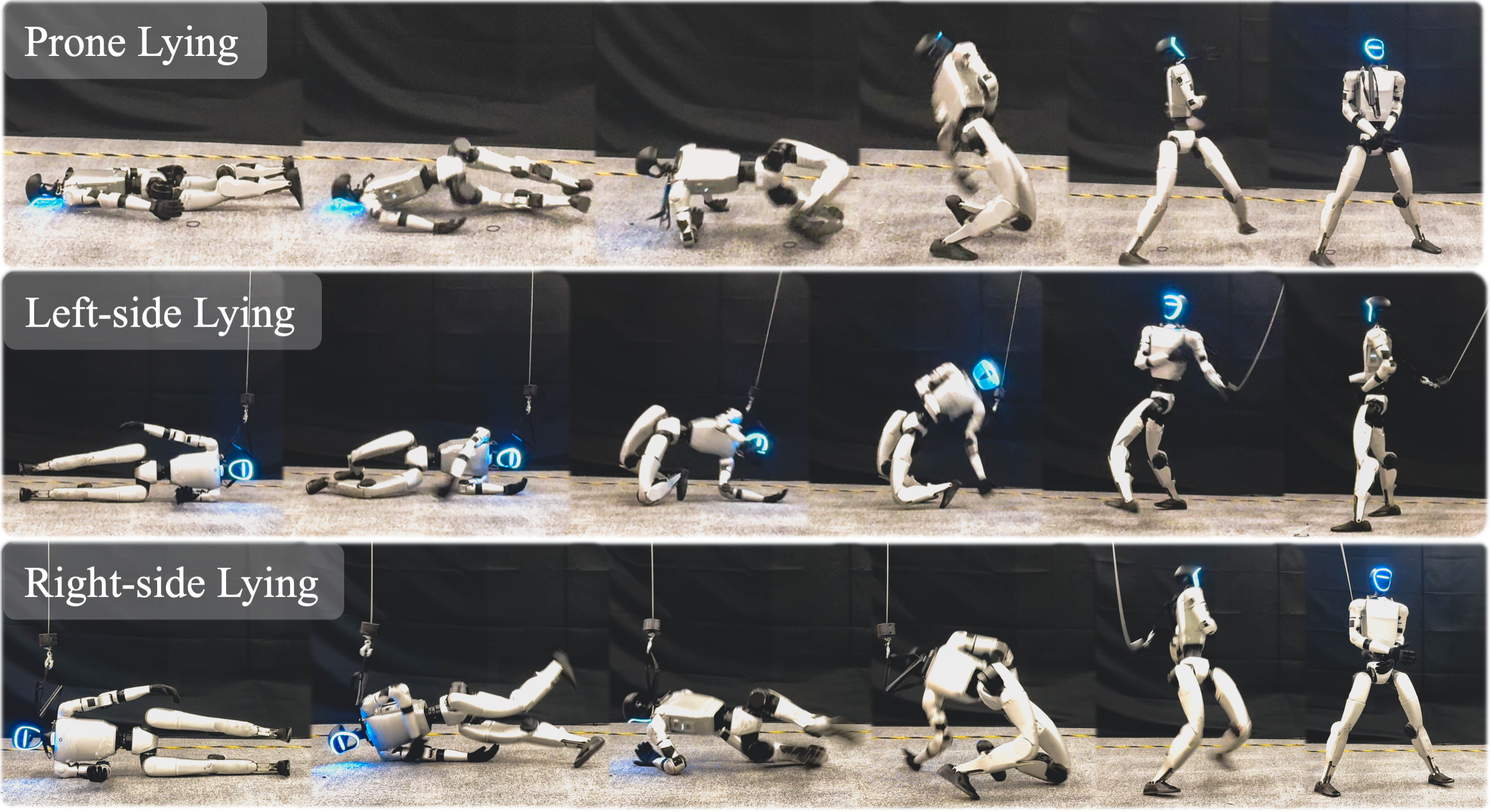}
    \caption{\textbf{More diverse postures.} HoST can learn across prone postures on the ground. The learn policies can also handle side-lying postures.}
    \vspace{-0.15in}
    \label{fig:more_diverse_postures}
\end{figure}

\paragraphbegin{Fall recovery (\cref{fig:real_robustness}b).} Our controllers also exhibited strong resilience in recovering from large external forces without fully falling down. This capability is vital for humanoid robots navigating unpredictable real-world scenarios with sudden impacts or balance shifts. Testing showed that, even under abrupt perturbations, the robots regained their upright posture, demonstrating the effectiveness of our control strategies in maintaining dynamic stability.

\paragraphbegin{Dynamic balance (\cref{fig:real_robustness}c).} We further tested our controllers on a 15$^\circ$ slippery slope, simulating challenging real-world conditions such as unstable surfaces. The controllers not only maintained stability on the incline but also adjusted posture and center of mass in real time to counteract the slippery conditions. These results highlight the adaptability and stability of our controllers, ensuring humanoid robots can operate safely on diverse and unpredictable terrains.

\paragraphbegin{Standing stability (\cref{fig:standing_stability}).} Our controllers demonstrate strong standing stability, effectively resisting external disturbances after successful standing up. This stability is beneficial for integrating our controllers into existing control systems.

\subsection{Prone and Side-lying Postures on the Ground}

We demonstrate that HoST is capable of learning across prone postures on the ground, as visualized in \cref{fig:more_diverse_postures}. Besides, the learned policies can also handle side-lying postures without any tuning. However, there are significant differences in motion patterns between
prone and supine postures. This somehow limits our method:
when training from posture postures, harder constraints on hip
joints are necessary to prevent violent motions, making the
feasibility of joint training from prone and supine postures
unclear currently.

\subsection{Extend HoST to Large-size Humanoid Robots}

\begin{figure}[t]
    \centering
    \includegraphics[width=1\linewidth]{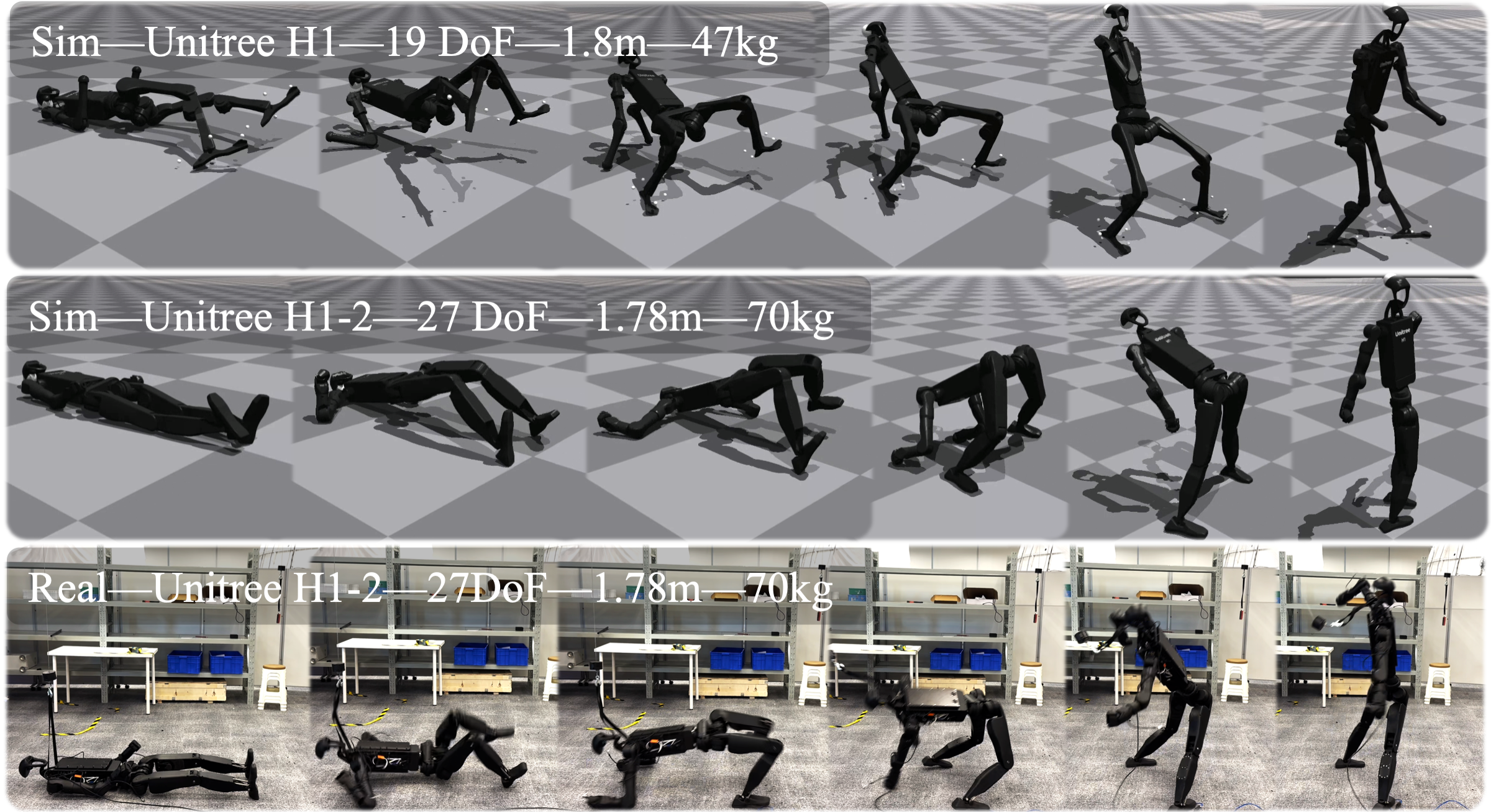}
    \caption{\textbf{Extension to large-size robots.} HoST can be easily extended to Unitree H1 and H1-2 humanoid robots with minor hyperparameter tuning.}
    \vspace{-0.15in}
    \label{fig:h1_h12}
\end{figure}

We believe that standing-up control is more challenging in
larger humanoid robots than in G1 due to their
increased weight and the limited actuators. As an initial test,
we extend HoST to Unitree H1  and H1-2. Simulation and real-world motions are visualized in \cref{fig:h1_h12}.
Compared to G1, we observe greater reliance on (i) upper-body contact with the ground and (ii) high hip actuation. While successful, two sim-to-real gaps emerge: (i) the need for high-stiffness joints to compensate for insufficient torques
and, which is consistent with the observation in \cref{subsec:sim2real}; (ii) noticeable deviations in upper-body posture. It remains unclear
whether these gaps originate from our framework or hardware
limitations. Identifying the source of these gaps is valuable in the future.

\section{Conclusion}
Our proposed framework, \ours, advances humanoid standing-up control by addressing the limitations of existing methods, which either neglect hardware constraints or rely on predefined motion trajectories. By leveraging reinforcement learning from scratch, \ours enables the learning of posture-adaptive standing-up motions across diverse terrains, ensuring effective sim-to-real transfer. The multi-critic architecture, along with smoothness regularization and implicit speed constraints, optimizes the controllers for real-world deployment. Experimental results with the Unitree G1 humanoid robot demonstrate smooth, stable, and robust standing-up motions in a variety of real-world scenarios. Looking forward, this work paves the way for integrating standing-up control into existing humanoid systems, with the potential of expanding their real-world applicability.

\section{Limitations and Future Directions}
While our method demonstrates strong real-world performance, we acknowledge several key limitations that should be addressed in the near future. 

\paragraphbegin{Perception of the environment.} Although proprioception alone is sufficient for many postures, some failures were observed during outdoor tests, such as standing from a seated position and colliding with surroundings. Integrating perceptual capabilities will help address this issue.

\paragraphbegin{More diverse postures.} We observe that training with both supine and prone postures has negatively impacted performance due to interference between sampled rollouts. Addressing this issue could further enhance capabilities like fall recovery and improve overall system generalization.

\paragraphbegin{Integration with existing humanoid systems.} Although this paper does not demonstrate integration with existing humanoid systems, we envision that standing-up control can be effectively incorporated into current humanoid frameworks to extend their real-world applications.

\section*{Acknowledgments}

This work is funded in part by the National Key R\&D Program of China (2022ZD0160201), and Shanghai Artificial Intelligence Laboratory. 

\bibliographystyle{plainnat}
\bibliography{references}
\appendix\label{app}
\subsection{More Experimental Details}\label{app:implementation}
\paragraphbegin{Hardware Setup.} We conducted our experiments using the Unitree G1 humanoid robot, which has a mass of 35 kg, a height of 1.32 m, and 23 actuated degrees of freedom (6 per leg, 5 per arm, and 1 in the waist). The robot is equipped with a Jetson Orin NX for onboard computation and uses an IMU and joint encoders to provide proprioceptive feedback.

\begin{table*}[t]
    \centering
    \caption{Reward functions and groups used for learning standing-up control. Reward functions within the same group are independently normalized, whose associated advantaged functions are estimated via a distinct critic. The bold symbols represent vectors. The $H$ with subscripts represents the threshold height of standing-up stages defined in \cref{subsec:multi_critic}. The $f_\mathrm{tol}$ is a gaussian-style function with a saturation bound, referring to~\cite{tassa2018deepmind,tao2022learning} for more details. 'G' denotes ground, and the letters in 'PSW' denote platform, slope, and wall, respectively.}
    \renewcommand{\arraystretch}{1.05} 
    \resizebox{1\linewidth}{!}{
    \begin{tabular}{l l l l}
    \toprule 
    Term & Expression & Weight & Description \\
    
    \midrule 
    \noalign{\vskip -0.2mm}
  \ourrow \textbf{(a) Task Reward}  & $r^\mathrm{task}$ & $w^\mathrm{task}=2.5$ & It specifies the high-level task objectives.\\ 
  \noalign{\vskip 0.4mm}\cdashline{1-4}\noalign{\vskip 0.8mm}
    Head height & $f_\mathrm{tol}\left(h_\mathrm{head},[1,\mathrm{inf}],1,0.1\right)$ & 1 & The head of robot head $h_\mathrm{head}$ in the world frame.\\
    Base orientation & $f_\mathrm{tol}\left(-{\theta}_{\mathrm{base}}^{\mathrm{z}},[0.99,\mathrm{inf}],1,0.05\right)$  & 1 & The orientation of the robot base represented by projected gravity vector.\\
    
    \cmidrule(r){1-4}
    \noalign{\vskip -0.2mm}
     \ourrow \textbf{(b) Style Reward}  & $r^\mathrm{style}$ & $w^\mathrm{style}=1$ & It specifies the style of standing-up motion.\\ 
  \noalign{\vskip 0.4mm}\cdashline{1-4}\noalign{\vskip 0.8mm}
    Waist yaw deviation & $\mathds{1}(| q_{\mathrm{waist}}| > 1.4)$  & $-10$ & It penalizes the large joint angle of the waist yaw. \\
    Hip roll/yaw deviation & $\mathds{1}( \max(|\bm{q}_{\mathrm{hip}}^{\mathrm{l,r}}|) > 1.4)\;|\;\mathds{1}( \min(|\bm{q}_{\mathrm{hip}}^{\mathrm{l,r}}|) > 0.9)$ & $-10$/$-10$ & It penalizes the large joint angle of hip roll/yaw joints. \\
    Knee deviation & $\mathds{1}( \max(\bm{q}_{\mathrm{knee}}^{\mathrm{l,r}}) > 2.85)\;|\;\mathds{1}( \min(\bm{q}_{\mathrm{knee}}^{\mathrm{l,r}}) < -0.06)$ & $\frac{-0.25 (G)}{-10 (PSW)}$ & It penalizes the large joint angle of knee joints. \\
    Shoulder roll deviation & $\mathds{1}( q_{\mathrm{shoulder}}^{l} < -0.02)\;|\;\mathds{1}( q_{\mathrm{shoulder}}^{r} > 0.02)$ & $-2.5$ & It penalizes the large joint angle of shoulder roll joint.  \\
    Foot displacement & $\exp\left(-2\times\|\bm{q}_{\mathrm{base}}^{\mathrm{xy}} - \bm{q}_{\mathrm{foot}}^{\mathrm{xy}}  \|^2.\mathrm{clip}(0.3,\mathrm{inf}) \right) 
    \times \mathds{1}(h_{\mathrm{base}} > H_{\mathrm{stage2}}) $ & $2.5/2.5$ & It encourages robot CoM locates in support polygon, inspired by~\cite{Goswami2004RateOC}.\\
    Ankle parallel & $(\mathrm{var}(\bm{q}_{\mathrm{left\;ankle}}^z) + \mathrm{var}(\bm{q}_{\mathrm{right\;ankle}}^z))/2 < 0.05$ & $20$ & It encourages the ankles to be parallel to the ground via ankle keypoints.\\
    Foot distance & $\|\bm{q}_{\mathrm{feet}}^l - \bm{q}_{\mathrm{feet}}^r \|^2 > 0.9$ & $-10$ & It penalizes a far distance between feet.\\
    Feet stumble & $\mathds{1}(\exists i,|\mathbf{F}_i^{\mathrm{x y}}|>3|F_i^\mathrm{z}|)$ & $\frac{0 (G)}{-25 (PSW)}$ & It penalizes a horizontal contact force with the environment. \\
    Shank orientation & $f_\mathrm{tol}(\mathrm
    {mean}(\mathrm{\bm{\theta}_{\mathrm{shank}}^{\mathrm{l,r}}[2]}),[0.8, \mathrm{inf}], 1, 0.1) \times \mathds{1}(h_{\mathrm{base}} > H_{\mathrm{stage1}}) $ & $10$ & It encourages the left/right shank to be perpendicular to the ground.\\
    Base angular velocity & $\mathrm{exp}(-2\times\|\bm{\omega}^\mathrm{xy}_\mathrm{base}\|^2) \times \mathds{1}(h_{\mathrm{base}} > H_{\mathrm{stage1}}) $ & $1$ & It encourages low angular velocity of the during rising up.  \\
    
    \cmidrule(r){1-4}
    \noalign{\vskip -0.2mm}
     \ourrow \textbf{(c) Regularization Reward}  & $r^\mathrm{regu}$ & $w^\mathrm{regu}=0.1$ & It specifies the regulariztaion on standing-up motion.\\ 
  \noalign{\vskip 0.4mm}\cdashline{1-4}\noalign{\vskip 0.8mm}
    Joint acceleration & $\|\ddot{p}\|^2$ & $-2.5 e^{-7}$ & It penalizes the high joint accelrations.\\
    Action rate & $\|a_t-{a}_{t-1}\|^2$ & $-1e^{-2}$ & It penalizes the high changing speed of action.\\
    Smoothness & $\|{a}_t-2 {a}_{t-1}+{a}_{t-2}\|^2$ & $-1e^{-2}$ & It penalizes the discrepancy between consecutive actions.  \\
    Torques & $\|\bm{\tau}\|^2$ & $-2.5e^{-6}$ & It penalizes the high joint torques.\\
    Joint power & $|\bm{\tau} \| \dot{p}|^{T}$ & $-2.5e^{-5}$ & It penalizes the high joint power\\
    Joint velocity & $ \|\dot{p}\|_{2}^{2}$ & $-1 e^{-4}$ & It penalizes the high joint velocity.\\
    Joint tracking error & $\|{p}_t - {p}^{\mathrm{target}}_{t}\|^{2}$ & $-2.5e^{-1}$ & It penalizes the error between PD target (\cref{eq:pd}) and actual joint position.  \\
    Joint position limits & $\sum_i[(p_i -p_i^{\mathrm{Lower}}).\mathrm{clip}(-\mathrm{inf},0) + (p_i -p_i^{\mathrm{Higher}}).\mathrm{clip}(0,\mathrm{inf}) ]$ & $-1e^2$ & It penalizes the joint position that beyond limits.\\
    Joint velocity limits & $\sum_i[(|\dot{p}_i| -\dot{p}_i^{\mathrm{Limit}}).\mathrm{clip(0,\mathrm{inf})}]$ & $-1$ & It penalizes the joint velocity that beyond limits. \\

    \cmidrule(r){1-4}
    \noalign{\vskip -0.2mm}
     \ourrow \textbf{(d) Post-task Reward}  & $r^\mathrm{post}$ & $w^\mathrm{post}=1$ & It specifies the desired behaviors after a successful standing up.\\ 
  \noalign{\vskip 0.4mm}\cdashline{1-4}\noalign{\vskip 0.8mm}
    Base angular velocity & $\mathrm{exp}(-2\times\|\bm{\omega}^\mathrm{xy}_\mathrm{base}\|^2) \times \mathds{1}(h_{\mathrm{base}} > H_{\mathrm{stage2}}) $ & $10$ & It encourages low angular velocity of robot base after standing up.  \\
    Base linear velocity & $\mathrm{exp}(-5\times\|\bm{v}^\mathrm{xy}_\mathrm{base}\|^2) \times \mathds{1}(h_{\mathrm{base}} > H_{\mathrm{stage2}}) $ & $10$ & It encourages low linear velocity of robot base after standing up.  \\
    Base orientation & $\exp(-5\times\|\bm{\theta}_{\mathrm{base}}^{\mathrm{xy}}\|^2 \times \mathds{1}(h_{\mathrm{base}} > H_{\mathrm{stage2}})$ & 10 & It encourages the robot base to be perpendicular to the ground. \\
    Base height & $\exp(-20\times\|{h}_{\mathrm{base}} - {h}_{\mathrm{base}}^{\mathrm{target}}\|^2 \times \mathds{1}(h_{\mathrm{base}} > H_{\mathrm{stage2}})$  & 10 & It encourages the robot base to reach a target height. \\
    Upper-body posture & $\exp(-0.1\times \| p_{\mathrm{upper}}-p_{\mathrm{upper}}^\mathrm{target} \|^2) \times \mathds{1}(h_{\mathrm{base}} > H_{\mathrm{stage2}})$ & 10 & It encourages the robot to track a target upper body postures. \\
    Feet parallel & $\exp(-20\times |h_{\mathrm{feet}}^l - h_{\mathrm{feet}}^r|.\mathrm{clip}(0.02, \mathrm{inf})) \times \mathds{1}(h_{\mathrm{base}} > H_{\mathrm{stage2}})$ & 2.5 & In encourages the feet to be parallel to each other.\\

    \bottomrule
    \end{tabular}}
    \vspace{-0.1in}
    \label{table:reward_functions} 
\end{table*}

\paragraphbegin{Evaluation Protocol.} Each policy is evaluated on each terrain with 5 repetitions of 250 episodes each, totaling 1250 episodes. We report the mean and standard deviation of performance. The target standing-up height is set to 0.6m for the slope terrain and 0.7m for all other terrains during evaluation.

\paragraphbegin{Robustness Test.} The CoM bias and sagittal force are set on the x-axis direction of the robot. The initial joint angle offset is applied to all joints of the robot. The random torque dropout is applied to each simulation step (200Hz), where the torques are set to zero if being dropout. 

\paragraphbegin{Expression of metrics:} Smoothness $E_{\mathrm{smth}}$ is computed via $\sum_{t=0}^{T-2} \|{p}_{t+2}-2 {p}_{t+1}+{p}_{t}\|^2$, where $p_t$ is the joint positions. Energy $E_{\mathrm{engy}}$ is computed via $\sum_{t=0}^{h_{\mathrm{base}}<H_{\mathrm{stage2}}} |\bm{\tau}_t|\cdot| \dot{p}_t|^{T}\mathrm{dt}$ approximately, where $\bm{\tau}_t$ is joint torques and $\dot{p}_t$ is joint velocities, $\mathrm{dt}$ (0.02s) is the time of a single policy step.

\subsection{More Implementation Details}\label{app:hyperparams}

\paragraphbegin{Curriculum Setup.} The curriculum adjustment condition is consistent for both the vertical force and action bound: the head height $h_{\mathrm{head}}$ must reach a target height $H_{\mathrm{head}}$ by the end of each episode. Initially, the vertical force $\mathcal{F}$ is set to 200N, and the action bound $\beta$ is set to 1. Upon reaching the target head height, the vertical force decreases by 20N, and the action bound decreases by 0.02. The lower bounds for the vertical force and action bound are 0N and 0.25, respectively.

\paragraphbegin{Stage Division.} The first stage involves righting the body, where we set $H_{\mathrm{stage1}}$ to 0.45m. The second stage involves rising the body, with $H_{\mathrm{stage2}}$ set to 0.65m.

\paragraphbegin{Reward Functions.} We present the complete set of reward functions and their detailed descriptions in \cref{table:reward_functions}. Several regularization reward terms are adapted from prior work~\cite{kumar2021rma,long2024learning,He2024OmniH2OUA}. Additionally, we incorporate a tolerance reward, $f_\mathrm{tol}(i, b, m, v)$, as defined in~\cite{tassa2018deepmind, tao2022learning}. This reward is computed as a function of an input value $i$, which is constrained by three parameters: bounds $b$, margin $m$, and value $v$. The bounds $b$ define the region where the reward is 1 if $i$ lies within the bounds. Outside this region, the reward smoothly decreases according to a Gaussian function, reaching the value $v$ at a distance determined by the margin $m$.

\begin{wrapfigure}{r}{0.1\textwidth}
  \centering
  \vspace{-0.16in}
  \includegraphics[width=0.1\textwidth]{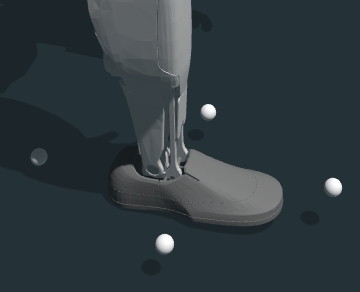}
  \label{fig:imi}
  \vspace{-0.3in}
\end{wrapfigure}
\paragraphbegin{Ankle parallel reward} is calculated as the variance of keypoints' height of the ankle, as visualized in the right figure. These keypoints are handcrafted without collision models. 

\paragraphbegin{PPO Implementation.} Our PPO implementation follows the framework outlined in~\cite{Rudin2021LearningTW}. The actor network consists of a 3-layer MLP with hidden dimensions [512, 256, 128], while each critic network is a 2-layer MLP with hidden dimensions [512, 256]. Each iteration includes 50 steps per environment, with 5 learning epochs and 4 mini-batches per epoch. The discount factor $\gamma$ is set to 0.99, the clip ratio is set to 0.2, and the entropy coefficient is 0.01. The multi-critic architecture is based on previous work~\cite{mysore2022multi}, where each advantage function is independently calculated and normalized within its corresponding reward group.

\paragraphbegin{Baseline Implementations.} \ours-w/o-MuC represents a baseline with a single value network, essentially a standard RL implementation. \ours-w/o-Force-RND removes the vertical force curriculum and introduces an RND reward with a coefficient of 0.2~\cite{burda2019exploration}. \ours-Bound0.25 uses a fixed action bound of $\beta = 0.25$ without a curriculum. \ours-w/p-$r^{\mathrm{style}}$ eliminates all style-related reward functions. 
Lastly, \ours-History modifies the history length of states while keeping other implementations unchanged.

\paragraphbegin{Terrains.} The heights of the platforms range from 20cm to 92cm. The slope inclination varies from approximately 1° to 14°. The wall inclination spans from approximately 14° to 84°.

\paragraphbegin{PD Controller.} In simulation, the stiffness values are set as 100 for the upper body, 40 for the ankle, 150 for the hip, and 200 for the knee. The damping values are set to 4 for the upper body, 2 for the ankle, 4 for the hip, and 6 for the knee. High stiffness values for the hip and knee are used due to the high torque demands during the standing-up process. During real-world deployment, we observe a significant torque gap between simulation and reality (see \cref{fig:sim2real}). Thus, the stiffness of the hip and knee are adjusted to 200 and 275, respectively.
\begingroup
\setlength{\tabcolsep}{8pt}
\begin{table}[h]
    \centering
    \vspace{-0.2in}
    \begin{tabular}{c cc c cc c cc}
    \toprule
  \multirow{2}{*}{Joint} &  \multicolumn{2}{c}{G1}  &  & \multicolumn{2}{c}{H1}  &  & \multicolumn{2}{c}{H1-2}\\ 
   \cmidrule{2-3} \cmidrule{5-6} \cmidrule{8-9}
   & Kp & Kd &  & Kp & Kd &  & Kp & Kd \\
    \midrule[0.8pt] 
    Hip & 150 & 4 & & 350 & 4 & & 350 & 4\\
    Knee & 200 & 6 & & 350 & 4 & & 350 & 4\\
    Ankle & 40 & 2 & & 120 & 2 & & 120 & 2\\
    Shoulder & 100 & 4 & & 350 & 4 & & 350 & 4\\
    Elbow & 100 & 4 & & 350 & 4 & & 350 & 4\\
    Waist & 100 & 4 & & 200 & 4 & & 200 & 4\\
    \bottomrule
    \end{tabular}
    \vspace{-0.1in}
\end{table}
\endgroup

\paragraphbegin{Observation noises} are without curriculum, set as below: 
\begin{table}[h]
    \centering
    \vspace{-0.1in}
    \setlength{\tabcolsep}{4pt}
    \resizebox{1\linewidth}{!}{
    \begin{tabular}{l|cc ccc}
    \toprule 
   Observation & Ang. Velocity & Pitch \& Roll & DoF Position & DoF Velocity & Action Rescaler  \\ 
    \midrule
    Noise Scale & $\mathcal{U}(-0.2,0.2)$ & $\mathcal{U}(-0.05,0.05)$ & $\mathcal{U}(-0.01,0.01)$ & $\mathcal{U}(-1.5,1.5)$ & $\mathcal{U}(-0.025, 0.025)$ \\
    \bottomrule
    \end{tabular}}
    \vspace{-0.1in}
    \label{table:comparision_method}
\end{table}

\paragraphbegin{Learning across prone postures.} We make the following adjustment to work the algorithm: more strict constraints on hip joint deviation rewards, weights for reward groups, and additional thigh orientation reward functions as a replacement for shank orientation rewards.  

\paragraphbegin{Extending HoST to Unitree H1 and H1-2.} We make the following adjustment to work the algorithm: scale of pulling force, height for curriculum, height for stage division, target postures, PD controllers, observation and action spaces. In our implementations, H1 has 19 actuators and H1-2 has 27 actuators. During the hardware deployment, the stiffness of hip and knee joints are amplified to 1.5 times than the simulation ones, similar to G1. We present more instructions in our code repository. Please refer to there for more details.

\end{document}